%% file: main.tex
\documentclass[nohyperref]{article}

\input{math_commands.tex}

\usepackage{microtype}
\usepackage{graphicx}
\usepackage{subfigure}
\usepackage{booktabs} 

\usepackage{hyperref}

\usepackage[accepted]{icml2022}

\usepackage{amsmath}
\usepackage{amssymb}
\usepackage{mathtools}
\usepackage{amsthm}

\usepackage[capitalize,noabbrev]{cleveref}

\theoremstyle{plain}

\theoremstyle{definition}

\theoremstyle{remark}

\usepackage[textsize=tiny]{todonotes}

\usepackage{caption}
\usepackage{bm}
\usepackage{color,graphicx}
\usepackage{wrapfig}
\usepackage{float}
\usepackage{multirow}
\usepackage{array}
\usepackage{booktabs}
\usepackage{gensymb}

\renewcommand{\baselinestretch}{.96}

\newcolumntype{P}[1]{>{\centering\arraybackslash}p{#1}}
\newcolumntype{M}[1]{>{\centering\arraybackslash}m{#1}}

\newboolean{include-notes}
\setboolean{include-notes}{false}

\newboolean{fff}
\setboolean{fff}{false}

\newcommand{\MicahComment}[1]{\ifthenelse{\boolean{include-notes}}
 {{\color{cyan}M: #1}}{}}
 
\newcommand{\henri}[1]{\ifthenelse{\boolean{include-notes}}
 {{\color{blue}AD: #1}}{}}

\newcommand{\adnote}[1]{\ifthenelse{\boolean{include-notes}}
 {{\color{blue}AD: #1}}{}}
 \newcommand{\adremove}[1]{\ifthenelse{\boolean{include-notes}}
 {{\color{blue}AD: \sout{#1}}}{}}
 
 \newcommand{\dhmnote}[1]{\ifthenelse{\boolean{include-notes}}
 {{\color{purple}DHM: #1}}{}}
 
  \newcommand{\lcnote}[1]{\ifthenelse{\boolean{include-notes}}
 {{\color{green}LC: #1}}{}}
 
\newcommand{\ar}[1]{\ifthenelse{\boolean{include-notes}}
 {{\color{purple}AR: #1}}{}}
 
 \newcommand{\sm}[1]{\ifthenelse{\boolean{include-notes}}
 {{\color{orange}SM: #1}}{}}

\newcommand{\Prob}[1]{\mathbb{P}(#1)}

\newcommand{\Expectation}[1]{\mathbb{E}[#1]}
\newcommand{\ExpectationWRT}[2]{\mathbb{E}_{#1}[#2]}

\newcommand{\evidence}{x}
\newcommand{\embedding}{x}
\newcommand{\is}{z}

\newcommand{\Pref}{U}
\newcommand{\pref}{u}

\newcommand{\rew}{\hat{r}}
\newcommand{\itemm}{x}
\newcommand{\choice}{x}
\newcommand{\slate}{s}
\newcommand{\slatebelief}[1]{{b^H_{#1}(s)}}

\newcommand{\prg}[1]{\noindent\textbf{#1}}

\icmltitlerunning{Estimating and Penalizing Preference Shifts Induced by RSs}

\begin{document}

\twocolumn[
\icmltitle{Estimating and Penalizing Induced Preference Shifts in Recommender Systems}



\icmlsetsymbol{equal}{*}

\begin{icmlauthorlist}
\icmlauthor{Micah Carroll}{ucb}
\icmlauthor{Anca Dragan}{ucb}
\icmlauthor{Stuart Russell}{ucb}
\icmlauthor{Dylan Hadfield-Menell}{mit}
\end{icmlauthorlist}

\icmlaffiliation{ucb}{Berkeley}
\icmlaffiliation{mit}{MIT}

\icmlcorrespondingauthor{}{mdc@berkeley.edu}

\icmlkeywords{Machine Learning, ICML}

\vskip 0.3in
]



\printAffiliationsAndNotice{}  

\begin{abstract}
The content that a recommender system (RS) shows to users influences them. Therefore, when choosing a recommender to deploy, one is implicitly also choosing to induce specific internal states in users. Even more, systems trained via long-horizon optimization will have direct incentives to manipulate users: in this work, we focus on the incentive to shift user \emph{preferences} so they are easier to satisfy. 
We argue that -- before deployment -- system designers should: \textit{estimate} the shifts a recommender would induce; \textit{evaluate} whether such shifts would be undesirable; and perhaps even \textit{actively optimize} to avoid problematic shifts. 
These steps involve two challenging ingredients: \emph{estimation} requires anticipating how hypothetical algorithms would influence user preferences if deployed -- we do this by using historical user interaction data to train a predictive user model which implicitly contains their preference dynamics;
\emph{evaluation} and \emph{optimization} additionally require metrics to assess whether such influences are manipulative or otherwise unwanted -- we use the notion of ``safe shifts'', that define a trust region within which behavior is safe: for instance, the natural way in which users would shift without interference from the system could be deemed ``safe''. 
In simulated experiments, we show that our learned preference dynamics model is effective in estimating user preferences and how they would respond to new recommenders. Additionally, we show that recommenders that optimize for staying in the trust region can avoid manipulative behaviors while still generating engagement.
\end{abstract}

\vspace{-2em}
\section{Introduction}

Recommender systems (RSs) generally show users feeds of items with the goal of maximizing their engagement, and users choose what to click on based on their preferences. Importantly, the recommender's actions are not independent of changes in users' internal states: simple changes in the content displayed to users can affect their behavior \citep{wilhelm_practical_2018, hohnhold_focusing_2015}, mood \citep{kramer_experimental_2014}, beliefs \cite{allcott_welfare_nodate}, and preferences \citep{adomavicius_recommender_2013, epstein_search_2015}.






\MicahComment{be more clear about differences with Deepmind paper and related work}

Given the dependence between users' internal state and the recommender system, when a system designer chooses a specific recommender algorithm (policy), they are implicitly also choosing how to influence user's behaviors, mood, preferences, etc. While traditionally RS policies have been \textit{myopic} (tended at satisfying users' current desires), optimizing \textit{long-term} user engagement has been a growing trend -- typically via reinforcement learning \citep{afsar_reinforcement_2021}; these non-myopic policies are commonly referred to as long-term value, or LTV, systems. However, these policies will have incentives to manipulate users as a side-effect \cite{albanie_unknowable_2017, krueger_hidden_2020, evans_user_2021}: for example, certain preferences are easier to satisfy than others, leading to more potential for engagement -- this could be because of availability of more content for some preferences compared to others, or because strong preferences for a particular type of content lead to higher engagement than more neutral ones. 
This can make LTV systems a particularly poor choice for avoiding undesired shifts in user's behaviors, moods, preferences, etc. While it has been proposed to prevent the RS from reasoning about manipulation pathways (e.g., by keeping it myopic) \citep{krueger_hidden_2020, farquhar_path-specific_2022}, even such systems can influence users in systematically undesirable ways \citep{jiang_degenerate_2019, mansoury_feedback_2020, chaney_how_2018}.

In this work, we focus on user's \textit{preference} shifts, which is a particularly challenging problem that has been gaining more attention \cite{franklin_recognising_2022}. Any recommender policy will have some influence on user preferences. While this might seem cause for concern, the degree to which undesirable and manipulative preference influence occurs in practice has yet to be measured. Moreover, no framework has yet been proposed to \emph{explicitly account for a policy's influence on users when choosing which policy to deploy}. We attempt to propose such a framework here -- requiring us to tackle two critical problems.

\begin{figure*}[h!]
    \vskip -0.8em
    \centering
    \includegraphics[width=0.9\textwidth]{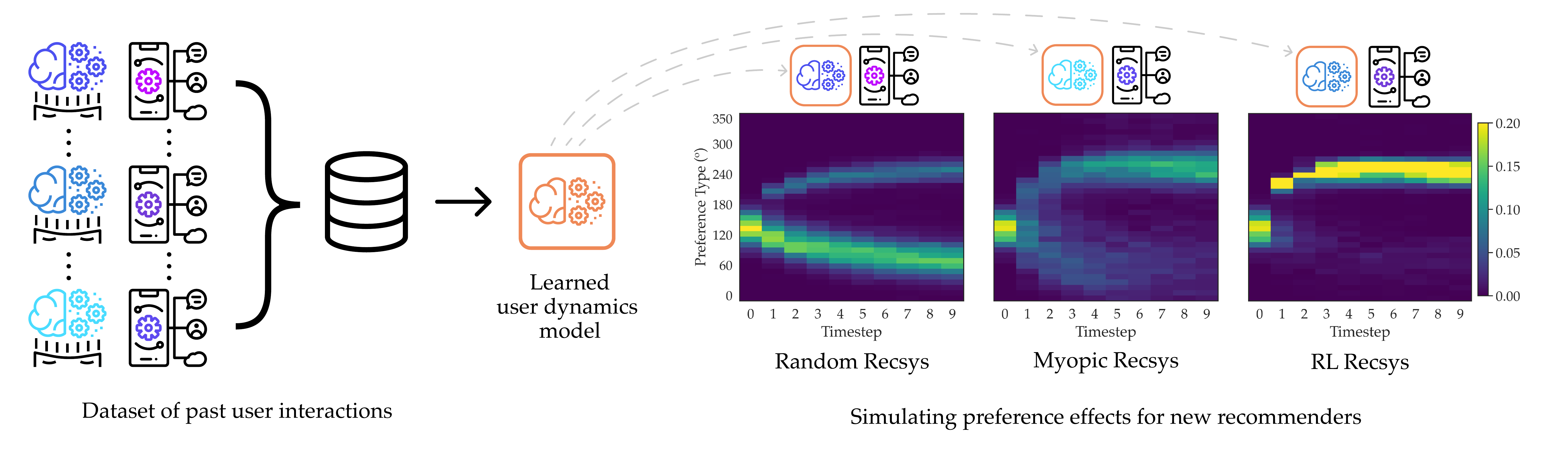}
    \vskip -0.19in
    \caption{\textbf{Proposal for anticipating user preference shifts.} Based on past user interactions, we can train a model of user preference dynamics, which can be used to anticipate how preferences will evolve under different recommenders. On the right we show the preferences induced by different RS policies across the same cohort of users. In our simulated environment, preferences (y-axis) are in 1D and change over time (x-axis) as users interact with the policy. We see that using RL drives user preferences to one spot. This is also true with a myopic policy, but to a lesser degree. These shifts are different from what we introduce as the ``natural'' evolution of user preferences.}
    \vskip -0.24in
    \label{fig:example}
\end{figure*}

Firstly, we need to anticipate what preference shifts a RS policy will cause \emph{before it is deployed}. This alone would enable system designers to generate visualizations such as those in \Figref{fig:example} (explained in more depth in our experiments): these could surface whether certain RS policies tend to induce shifts that are significantly different from others, and allow to assess whether such shifts seem manipulative or otherwise unwanted. We show how to do this using historical user interaction data: we train a model that implicitly captures preference dynamics, which can
anticipate how user preferences would evolve under a new policy.

The second critical problem is having quantitative metrics for evaluating whether an evolution of preferences is actually unwanted: a computable metric not only simplifies evaluation, but could also enable recommenders to be optimized against unwanted shifts. The issue arises from the fact that \textit{standard engagement metrics are preference-change-agnostic}: they do not directly assign value to preference shifts (or shifts in any other aspect of the user's state, for that matter). Even if a system were to completely overwrite a user’s preferences, as long as the user is engaged, the system would be ``performing well''. We thus introduce some preliminary metrics which assign value to preference changes -- based on how the users' preferences would have changed in the absence of the recommender -- in an attempt to isolate the recommender's influence over the user's shifts. 

For such metrics, instead of defining desirable or undesirable shifts directly, we provide a framework for conservatively defining ``safe shifts'': we non-exhaustively list certain shifts that we trust not to be particularly problematic, and measure the extent to which other shifts deviate from them. If the shifts induced by a policy differ significantly from the safe ones, they should be flagged as warranting more investigation.
A candidate for safe shifts that we introduce is how user preferences would shift if they were interacting with a random recommender -- which we call ``natural preference shifts''. One can think of this as an approximation to not having a recommender system at all.
\Figref{fig:example} (left) shows this natural preference evolution for our running example, and how user preferences stay somewhat diffuse but drift towards the opposite mode that the RL and myopic policies push them to.
Note that while our metrics can effectively penalize undersired shifts, it comes at a cost: natural shifts, and in fact any lists of safe shifts that we define, are unlikely to be exhaustive, which means the approach will conservatively penalize policies that might in reality be safe.

To demonstrate both the estimation of preference shifts and their evaluation, we set up a testing environment in which we emulate ground truth user behavior by drawing from a model of preference shift from prior work \citep{bernheim2021theory}. 
We first show that our learned preference shift estimation model -- trained using historical user interaction data -- can correctly anticipate user preference shifts almost as well as knowing the true preference dynamics. Additionally, we show qualitatively that in this environment, RL and even myopic recommenders lead to potentially undesired shifts. Further, we find that our evaluation metric can correctly flag which policies will produce undesired shifts, and evaluates the RL policy from \Figref{fig:example} as 35\% worse than the myopic one, which is in turn 40\% worse than our policy which is penalized for manipulating user's preferences. Our results also suggest that evaluating our metric using the trained estimation model correlates to using ground truth preference dynamics, and that optimizing for safe shifts does lead to higher scoring (more safe) policies.

Although this work just scratches the surface of finding the right metrics for unwanted preference shifts and evaluating them in real systems, our results already have implications for the development of recommender systems: in order to ethically use recommenders at scale, we must take active steps to \textit{measure} and \textit{penalize} how such systems shift users' internal states. In turn, we offer a roadmap for how one might be able to do so by learning from user interaction data, and put forward a framework for specifying ``safe shifts'' for detecting and controlling such incentives.


\section{Related Work} 

\prg{RS effects on users' internal states.} A variety of previous work considers how RSs algorithms might affect users: influencing user's preferences for e-commerce purposes \citep{haubl_preference_2003, cosley_is_2003, gretzel_persuasion_2006}, altering people's moods for psychology research \citep{kramer_experimental_2014}, ``nudging'' users' opinions or behaviors \citep{jesse_digital_2021, matz_psychological_2017, weinmann_digital_2015}, exacerbate \citep{hasan_excessive_2018} cases of addiction to social media \citep{andreassen_online_2015}, or increase polarization \citep{stray_designing_2021}. 
There have been three main types of approaches to quantitatively estimating \textit{effects of RSs' policies} on users: 1) analyzing static datasets of interactions directly \citep{nguyen_exploring_2014, ribeiro_auditing_2019, juneja_auditing_2021, li_modeling_2014},
2) simulating interactions between users and RSs based on hand-crafted models of user dynamics \citep{chaney_how_2018, bountouridis_siren_2019, jiang_degenerate_2019, mansoury_feedback_2020, yao_measuring_2021, ie_recsim_2019}, or 3) using access to real users and estimating effects through direct interventions \citep{holtz_engagement-diversity_2020, matz_psychological_2017}.
We see our approach as an improvement on 2), in that we propose to implicitly \textit{learn} user dynamics instead of hand-specifying them. While we still assume a known choice model (how users choose content based on their preferences), such assumption is much less restrictive than assuming fully known dynamics. Our approach is most similar to \citet{hazrati_recommender_2022}, but focused on preferences rather than behavior.


\prg{Neural networks for recommendation and human modeling.} While data-driven models of human behavior have been used widely in RSs as click predictors \citep{zhang_deep_2019, covington_deep_2016, cheng_wide_2016, okura_embedding-based_2017, mudigere_high-performance_2021, zhang_sequential_2014, wu2017recurrent} and for simulating behavior in the context RL RSs' training \citep{chen_generative_2019, zhao_toward_2019, bai_model-based_2020, shi_virtual-taobao_2018}, to our knowledge they have not been previously used for explicitly simulating and quantifying the \textit{effect on users} of hypothetical recommenders. 
We emphasize how human models can also be used as simulation mechanisms by anticipating RSs' policies impact on users' \textit{behaviors} and, as enabled by our method, even \textit{preferences}.

\prg{RL for RS.} Using RL to train RSs has recently seen a dramatic increase of interest \citep{afsar_reinforcement_2021}, with some notable work led by YouTube, Meta, and others \citep{ie_reinforcement_2019,  chen_top-k_2020, gauci_horizon_2019, cai_constrained_2022} -- leading to significant real-world performance increases.

\prg{Side effects and safe shifts.} Our work starts from a similar question to that of the side effects literature \citep{krakovna_penalizing_2019, kumar_realab_2020}, applied to preference change: given that the reward function will not fully account for the cost (or value) of preference shifts, how do we prevent undesired preference-shift side effects? Our notion of safe shifts corresponds to the choice of baseline in this literature -- see \citet{farquhar_path-specific_2022} for more information.


\section{Preliminaries} \label{sec:preliminaries}

\prg{Setup.} We model users as having time-indexed preferences $\pref_t \in \R^d$, which assign a scalar value to every possible item of content $\choice \in \R^d$: the value to the user derived from the engagement with item $\choice_t$ under $\pref_t$ is modeled as being given by $\rew_t(\pref_t) = \pref_t^T \choice_t$. The user's preferences $\pref_t$ together with other variables constitute the user's \textit{internal state} $\is_t$, which comprises a sufficient statistic for their long-term behavior, but which our method will not explicitly model. At every time step the user sees a slate $\slate_t$ (a list of items) produced by a recommender policy $\pi$, and chooses an item $\choice_t$. The policy $\pi$ maps history of slates and choices so far, $\slate_{0:t}, \choice_{0:t}$, to each new slate $\slate_{t+1}$: that is $\slate_{t+1} \sim \pi(\slate_{0:t}, \choice_{0:t})$. Upon making a choice, the user's internal state updates to $\is_{t+1}$. 

\prg{User choice model.} We are interested in inferring and predicting preferences, which we do not observe directly. As such, we make an assumption -- justified in \Appref{appendix:choice_model} -- about how user behavior (the user's choice $x_t$) relates to their current preferences $u_t$: that is, we assume the form of $\Prob{\itemm_t = \evidence | \slate_t, \pref_t}$ is known. We also test how poorly our method would perform if such a model were mis-specified.


\prg{Preference evolution as an NHMM.} The dynamics of the user's time-indexed internal state $\is_t$ -- which we don't assume to be known -- depend on the previous state $\is_{t-1}$ and on the last choice of slate by the recommender $\slate_t$ (which in turn depends on history because the policy $\pi$ uses history). This makes our setup a Non-homogeneous HMM (NHMM) \citep{hughes_non-homogeneous_1999, rabiner_introduction_nodate}, with hidden state $\is_t$ and time-dependent dynamics $\Prob{\is_{t+1}|\is_t, \slate_t}$ (the slate $\slate_t$ chosen by $\pi$ will affect the next internal state). See \Appref{appendix:nhmm} for more information. We will be using the NHMM inferences as an oracle benchmark for performance.

\section{Anticipating Users' Preferences Changes} \label{sec:future_estimation}

Our proposal boils down to the following: before deploying a new recommender policy $\pi'$, we first need to anticipate how that policy would change preferences and behavior of users. We will estimate these changes by predicting internal state evolution for each user in the population with whom the RS has already interacted, and aggregating these estimates to get an overall trend.  
More formally, for a set of $N$ users, we assume access to historical data of their interaction with a RS policy $\pi$: $\mathcal{D}_j=\{ \slate^\pi_{0:T}, \evidence^{\pi}_{0:T}\}$  for every user $j$. 
For each user, we are interested in estimating 
how their preferences would evolve if further interactions occurred with a new policy $\pi'$, which we denote $\pref^{\pi'}_{H}$ (where $H>T$).

\subsection{Estimation under known user dynamics}\label{subsub:fut_est_known}
We first describe how one could solve this problem optimally if one had oracle access to the true user internal state dynamics -- when such dynamics are unknown, we will try to approximate the inference process below while simultaneously learning the dynamics.

Assuming the user internal state dynamics $\Prob{\is_{t+1}|\is_t, \slate_t}$ to be known, our goal -- as mentioned above -- is to estimate what their preferences would be at a future timestep $H$ if policy $\pi'$ were to be deployed going forward, that is, $\pref_{H}^{\pi'}$ (for an example of the form of $\is_t$, see \Figref{fig:human_model}).
Given the assumption of known internal state dynamics, every component of the NHMM is known, meaning that we can perform exact inference over $\pref_{H}^{\pi'}$.
We can do so via a simple extension of HMM prediction: using the history of interactions $(\slate^\pi_{0:T}, \evidence^{\pi}_{0:T})$ to estimate $\Prob{\is^{\pi'}_{H} | \slate^\pi_{0:T}, \evidence^{\pi}_{0:T}}$ as shown in \Appref{appendix:future_est}.
As the preferences $\pref^{\pi'}_H$ are just one component of the internal state $\is_H^{\pi'}$, one can trivially recover them. 

\begin{figure}[t]
  \vspace{-0.8em}
  \centering
  \includegraphics[width=1\linewidth]{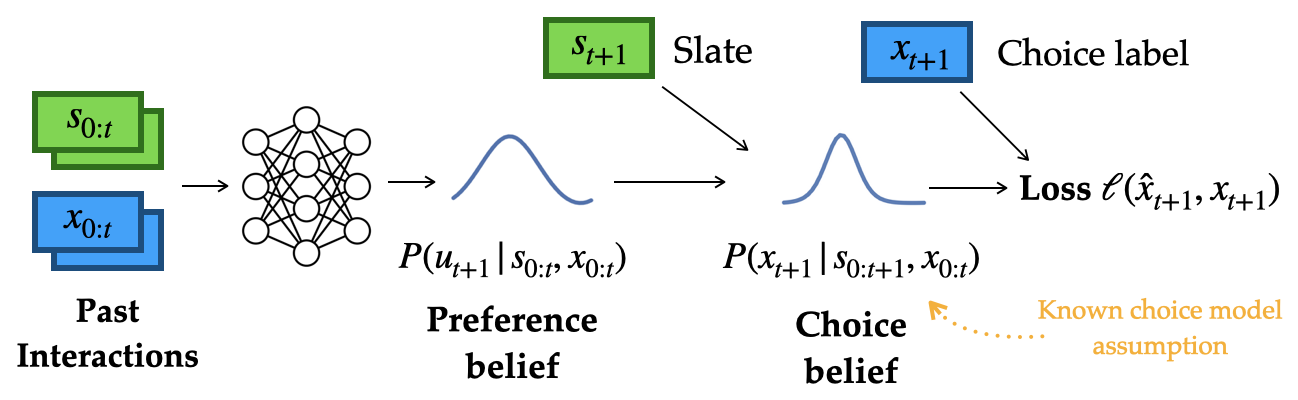}
  \vspace{-2.1em}
  \caption{\textbf{Future preference estimation model.} Given slates and choices up until the current timestep $\slate_{0:t}, \choice_{0:t}$, we train a network to predict beliefs over the next preferences $\pref_{t+1}$ which -- together with the next slate $\slate_{t+1}$ -- induce a distribution over content choices through the choice model. We can supervise the training with the actual choices users made for slates -- the network will learn to output preference beliefs which induce similar choices to those in the data.
  }
  \label{fig:choice_model}
  \vspace{-1.8em}
\end{figure}

\subsection{Estimation under unknown user dynamics}\label{subsub:fut_est_unknown}
In practice, one will not have access to an explicit representation of the user's internal state $z_t$, let alone its dynamics. We thus attempt to approximate the NHMM estimation task we are interested in by implicitly learning user preference dynamics from their past interaction data. Given a dataset of interactions over an horizon $T$ consisting of slates $\slate_{0:T}$ and corresponding choices $\choice_{0:T}$, we train a model to predict a next choice $x_{t+1}$ given a slate $s_{t+1}$ and the history $(s_{0:t},x_{0:t})$. What we are really interested in is the dynamics of preferences, so we structure the model as in \Figref{fig:choice_model}: output a distribution over preferences $u_{t+1}$, that, when used to make a choice over $s_{t+1}$ using our assumed choice model from \Secref{sec:preliminaries}, produces the observed choice $x_{t+1}$ with high likelihood.
Specifically, the model will output a belief over the next-timestep preferences $\Prob{\pref_{t+1} | \slate^\pi_{0:t}, \choice^\pi_{0:t}}$. Using the known choice model assumption from \Secref{sec:preliminaries}, we map the belief over preferences, together with the new slate $\slate_{t+1}$, to a distribution over content items: $\Prob{\choice_{t+1} | \slate_{0:t+1}, \choice_{0:t}} = \sum_{u_{t+1}} \Prob{\choice_{t+1} | \pref_{t+1}, \slate_{t+1}} \Prob{\pref_{t+1} | \slate_{0:t}, \choice_{0:t}}$.\MicahComment{maybe put this in appendix}

We now have a model which has learned to map histories of slates $\slate_{0:t}$ and choices $\choice_{0:t}$ to next-timestep preferences. This has been trained based on histories generated by interaction with a previous policy $\pi$ (or, more realistically, a set of previous policies) up to a time T, and we will use it to make predictions of user preferences that would be induced by further interacting with a new policy $\pi'$ starting at time T: $\Prob{\pref^{\pi'}_{H} | \slate^\pi_{0:T}, \choice^\pi_{0:T}}$ where $H > T$. 
This would require the predictor to generalize to new histories that would be induced by $\pi'$ after time T, never seen at training time. Nonetheless, we can be hopeful if the real-world dataset of interactions used for training is collected under many, diverse policies $\pi$ (which would likely be the case in practice).

\begin{figure*}[t]
  \vspace{-0.91em}
  \centering
  \includegraphics[width=\linewidth]{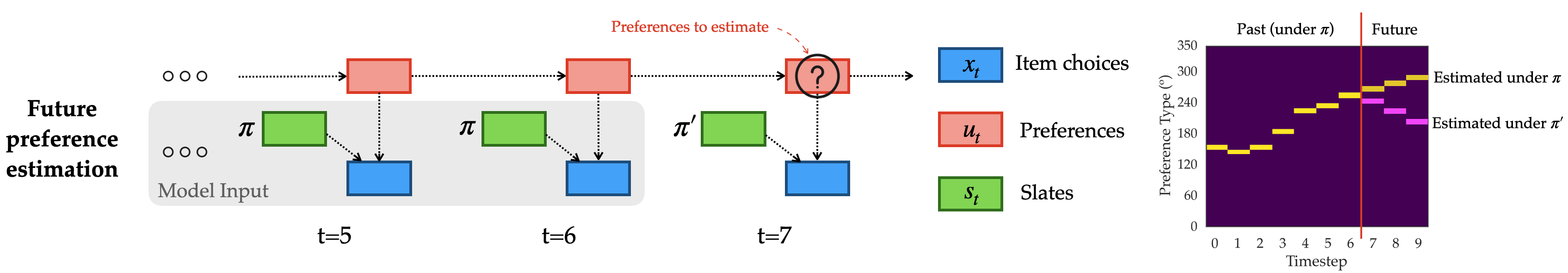}
  \vspace{-2.3em}
  \caption{\textbf{Simulating future user preferences and choices.} By iteratively using the future preferences estimation model (\Figref{fig:choice_model}) by inputting the observables (shaded, e.g. $\choice^\pi_{0:6}, \slate^\pi_{0:6}$), one can simulate how an existing user's preferences would evolve \textit{in the future} if they interacted with same policy $\pi$ or a different policy $\pi'$ (by selecting future slates with $\pi'$).}
  \label{fig:future_pref_est}
  \vspace{-1.5em}
\end{figure*}

To obtain a belief over $\pref^{\pi'}_{H}$ for a user $j$, we can sample \textit{simulated user preference trajectories} from the model to obtain a Monte Carlo estimate of the desired distribution -- as shown in \Figref{fig:future_pref_est}. We use the data $\mathcal{D}_j=\{ \slate^\pi_{0:T}, \evidence^{\pi}_{0:T}\}$ as input to the model to obtain a belief over the next-timestep preferences $u_{T+1}$. We then use such belief and a slate $\slate^{\pi'}_{T+1}$ sampled from policy $\pi'$ to simulate the user's choice $\choice^{\pi'}_{T+1}$. Treating this extra (simulated) step of interaction history for the user (the slate $\slate^{\pi'}_{T+1}$ and choice $\choice^{\pi'}_{T+1}$) as part of the observed history so far, we repeat this process to obtain preference estimates under $\pi'$ for future timesteps. By simulating multiple future trajectories, we can obtain beliefs over the expected preferences a user would have at any future timestep. See \Algref{alg:future_prediction} for the exact procedure. In \Appref{appendix:future_est} we show that this procedure does in fact approximate the desired estimate.



\section{Evaluating Preferences and Avoiding Unwanted Shifts} \label{sec:proxies}

Given the estimates of preference shifts under different policies that \Secref{sec:future_estimation} enables us to compute, it would be useful to have quantitative metrics for whether they are undesirable: this would enable to automatically evaluate policies and even actively optimize to avoid such shifts. To obviate the difficulty of explicitly defining unwanted shifts, we limit ourselves to defining some shifts that we somewhat trust -- ``safe shifts'' -- and flag preference shifts that largely differ from these safe shifts as potentially unwanted. This is similar to defining a region of the space which we trust \textit{in a risk-averse manner}. The underlying philosophy is that ``we don't know what good shifts are, but as long as you stay close to [family of safe shifts], things shouldn't go too bad''. 
To do this, we need both a notion of shifts we trust (``safe shifts''), and a notion of distance between different evolutions of preferences.

\prg{Notation.} We denote the $\pi$-induced engagement under ``safe-shift preferences'' $\pref_t^{\text{safe}}$ as $\rew_t(\pref_t^{\text{safe}}, \pi) = (\choice_t^{\pi})^T \pref^{\text{safe}}_t$. This means that $\pi$ was used to select the slate from which the user picked the item $\choice_t^\pi$, but $\choice_t^\pi$'s engagement is considered relative to different preferences $\pref_t^{\text{safe}}$ than those the user would have developed under $\pi$. 

\prg{Distance between shifts.} Consider a known preference-trajectory $\pref_{0:T}^\pi$ induced by $\pi$. We choose a metric between shifts such that engagement for the items $\choice_{0:T}^{\pi}$ chosen under policy $\pi$ is also high under the preferences one would have had under safe shifts $\pref_{0:T}^{\text{safe}}$.
We operationalize this notion of distance between $\pi$-induced shifts and safe shifts $\pref_{0:T}^{\text{safe}}$ as $D(\pref_{0:T}^{\pi}, \pref_{0:T}^{\text{safe}}) = \sum_t  \mathbb{E}\big[ \rew_t(\pref_t^{\pi}, \pi) - \rew_t(\pref_t^{\text{safe}}, \pi) \big] = \sum_t  \mathbb{E}_{\choice_t^\pi}\big[ (\choice_t^{\pi})^T \pref^{\pi}_t - (\choice_t^{\pi})^T \pref^{\text{safe}}_t \big]$. However, this operationalization could easily be substituted with others. In the case that safe shifts are random variables (a family of safe shifts), we consider the expected distance $\mathbb{E}_{\Pref_{0:T}^{\text{safe}}}\big[D(\pref_{0:T}^{\pi}, \Pref_{0:T}^{\text{safe}})\big]$.

\prg{Safe shifts: initial preferences.} As a first (very crude) proposal for safe shifts, we can consider using the initial preferences $\pref_0$ as our safe shifts $\pref_{0:T}^\text{safe}$: any deviation from the initial preferences $\pref_0$ will be flagged as potentially problematic with $D(\pref_{0:T}^{\pi}, \pref_{0})$.

\prg{Safe shifts: natural preference shifts (NPS).} One limitation of the above metric is that not all preference shifts are unwanted: people routinely change their preferences and behavior ``naturally''. But what does ``naturally'' even mean? We propose an idealized notion of natural shifts, by asking how preferences would evolve if the user were ``omniscient'' and have full agency over their preference evolution process, i.e. if would have access to all content directly and had the ability to process it, unhindered by a small and biased slate offered by the RS.
Unfortunately this is impractical, as we can never get data from such hypothetical users that can attend to all content when choosing what to consume. As an approximation, we consider \emph{random} slates, which at least eliminates the agency of the RS policy (which in turn can change the user's belief about the distribution of available content).
We therefore operationalize ``natural preferences'' $\pref^{\pi_{\text{rnd}}}$ as the preferences which the user would have interacting with a random recommender $\pi^\text{rnd}$, and we use $\mathbb{E}_{\Pref_{0:T}^{\text{rnd}}}\big[D(\pref_{0:T}^{\pi}, \Pref_{0:T}^{\pi_{\text{rnd}}})\big]$ as the metric.


\prg{Using safe shift metrics.} For how one can estimate such metrics, see \Appref{appendix:metrics_computation}. Once computed, they can be used both for evaluating preference shifts or for recommender optimization (as alternate proxies for ``value'' relative to simple engagement). However, clearly such metrics also fall short of fully capturing ``value'': we don’t want a system that actively tries to keep the user preferences static (as would result from blindly optimizing $D(\pref_{0:T}^{\pi}, \pref_{0})$). 

\prg{Penalized objective.} By considering a weighted sum of these metrics (as we will do in the penalized RL objective below), we try to lead the system to perform well under a variety of relatively-reasonable definitions of value, some of which are not preference-shift-agnostic. One can think of this as hedging our bets as to what the value (or cost) of induced preference shifts should be, and making explicit that it should not be zero, as current systems assume.

\prg{Penalized RS training.} By using the method from \Secref{sec:future_estimation}, one obtains a human model which can be used to \textit{simulate human choices} (in addition to preferences). Similarly to previous work, we can use this human model as a simulator for RL training of recommender systems \citep{chen_generative_2019, 
zhao_toward_2019, 
bai_model-based_2020}. 
During training, we can compute the metrics defined above penalize the current policy for causing any shifts that we have not explicitly identified as ``safe'' -- in what can be considered a ``risk-averse'' design choice.
We incorporate these metrics in the training of a new policy $\pi$ by adding the two distance metrics from above to the basic objective $\mathbb{E}\big[\rew_t(\pref_t^{\pi}) \big]$ of maximizing long-term engagement, leading to the updated objective $\mathbb{E}\big[\rew_t(\pref_t^{\pi}) + \nu'_1 \ \rew_t(\pref_0, \pi) + \nu'_2 \ \rew_t(\pref_t^{\pi_{\text{rnd}}}, \pi) \big]$ where $\nu'_1, \nu'_2$ are hyperparameters. This objective can be optimized either myopically or via long-horizon (RL) optimization. See \Appref{appendix:RL} for more details. \MicahComment{expectations over?}

\section{Experimental Setup} \label{sec:ex_setup}

\begin{figure*}[t]
    \vspace{-0.8em}
    \centering
    \includegraphics[width=0.75\linewidth]{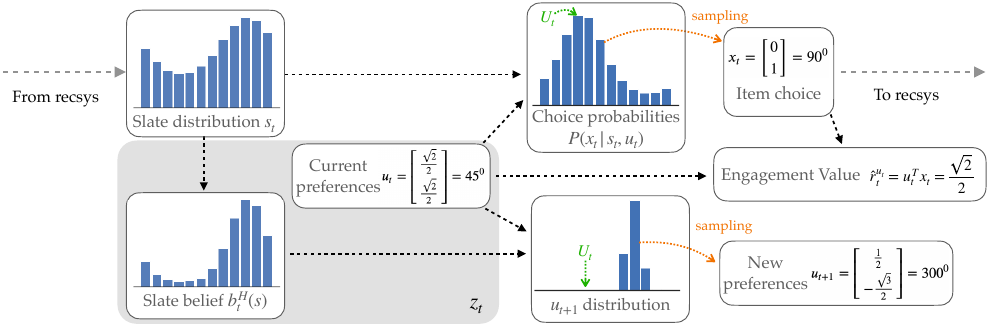}
    \vspace{-1em}
    \caption{\textbf{Ground truth human dynamics.} At each timestep, the user will receive a slate $\slate_t$. Given the user's preferences $\pref_t$, the slate $\slate_t$ induces a distribution over item choices $\Prob{\choice_t|\slate_t, \pref_t}$ from which the user samples an item $\itemm_t$ and receives an engagement value $\rew_t$ (unobserved by the RS). Additionally, $\slate_t$ induces a belief over the future slates in the user $\slatebelief{t}$. In turn $\slatebelief{t}$ -- together with $\pref_t$ -- form $z_t$ and induce a distribution over the next-timestep user preferences $\pref_{t+1}$.}
    \label{fig:human_model}
    \vspace{-1.5em}
\end{figure*}

\prg{Why simulation?} 
To test our method, we need to evaluate both recommenders which interact with users (rendering static datasets of user interaction unsuitable), as well as test our evaluation metrics themselves -- which are defined based on internal preferences (for which we never get ground truth in real interactions). We thus create a testing environment in which we can emulate user behavior and have access to their ground truth preferences.
Like previous work \citep{chaney_how_2018, bountouridis_siren_2019, jiang_degenerate_2019, mansoury_feedback_2020, yao_measuring_2021}, we simulate both a recommendation environment and human behavior. However, unlike such approaches, we use simulated human behavior for evaluation purposes only: our human model is learned exclusively from data that would be observed in a real-world RS (slates and choices), i.e. data of users (in our case the simulated users) interacting with previous RS policies -- meaning our approach could be applied to real user data of this form. A fundamental advantage of testing our method in a simulated environment is that \textit{we can actually evaluate how well our model is able to recover the preferences of our ``ground truth'' users, giving us insights about how our methods could perform with real users}.

\prg{Ground truth user dynamics.} See \Figref{fig:human_model} for a summary of the ground truth user dynamics we use for testing our method (and \Figref{fig:content} for more info on our environment setup). 
Following prior work \citep{ie_reinforcement_2019, chen_generative_2019} we assume that users choose items in proportion to (the exponentiated) engagement under their current preferences, i.e. that $\Prob{\itemm_t = \evidence | \slate_t, \pref_t}$ is given by the conditional logit model. In our setup, this reduces to
$\Prob{\itemm_t = \evidence | \slate_t, \pref_t} \propto \Prob{\slate_t = \itemm} e^{\beta_c \itemm^T \pref_t }$, with an additional term $\Prob{\slate_t = \choice}$ which takes into account how prevalent each item is in the slate -- and we assume such choice model to be also known by our method.
We adapt \cite{bernheim2021theory} to be our ground truth human preference dynamics. On a high-level, at each timestep users choose their next-timestep preferences to be more ``convenient'' ones -- trading off between choosing preferences that they expect will lead them to higher engagement and maintaining engagement under current preferences. See \Appref{appendix:human_model} for a proof of the logit model reduction and more details.

\prg{Simulated environment setup.} For ease of interpretation of the results, in our experiments we only consider a cohort of users whose initial preferences are concentrated around preference $\pref=130\degree$. To showcase preference-manipulation incentives to make users more predictable, we make the choice-stochasticity temperature $\beta_c$ a function of part of preference space one is in, with local optima $\beta_c(80\degree) = 1$ and $\beta_c(270\degree) = 4$. This causes these areas in preference space to be attractor points, as RSs are able to lead to higher engagement when users act less stochastically (see \Appref{appendix:human_model} for more details).

\prg{Dataset.} For all our experiments, we use a dataset of 10k trajectories (each of length 10), collected under a mixture of policies described in \Appref{appendix:datasets}. 7.5k trajectories are used for training our models and 2.5k for computing the validation losses and accuracies reported in \Secref{subsec:results_estimation}.

\prg{Training human models and penalized RS policies.} 
For training our human models, we use a bidirectional transformer model similar to that of \citet{sun_bert4rec_2019}, and only assume access to a dataset of historical interactions $\slate_{0:T}, \choice_{0:T}$. We train myopic and RL RS policies $\pi'$ using PPO \citep{PPO, liang_rllib_2018} and restrict the action space to 6 possible slate distributions for ease of training. 
For penalized training we give the three metrics $\rew_t(\pref_t^\pi)$, $\rew_t(\pref_0, \pi')$, $\rew_t(\pref_t^{\pi_{\text{rnd}}}, \pi')$ equal weight. For more details on human model and RL training, see respectively \Appref{appendix:hm_training} and \Appref{appendix:RL}.

\begin{figure}[t]
    \vskip -0.5em
    \centering
      \includegraphics[width=1\linewidth]{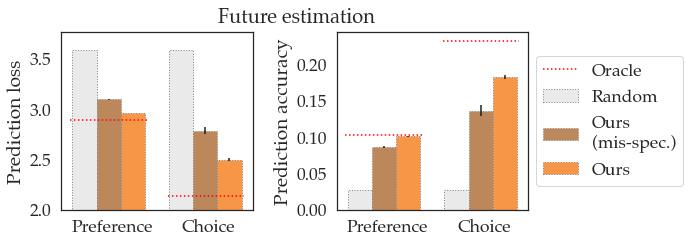}
      \vskip -1em
      \captionof{figure}{Validation losses and accuracies on held-out trajectories for the preference prediction task, averaged across timesteps. Both under the correct choice model and with some mis-specification, preference prediction performs similarly to oracle NHMM estimation (which additionally has access to the preference dynamics).}
      \label{fig:future_estimation}
    \vskip -0.5em
\end{figure}

\begin{figure}[t]
    \vskip -0.5em
    \centering
      \includegraphics[width=1\linewidth]{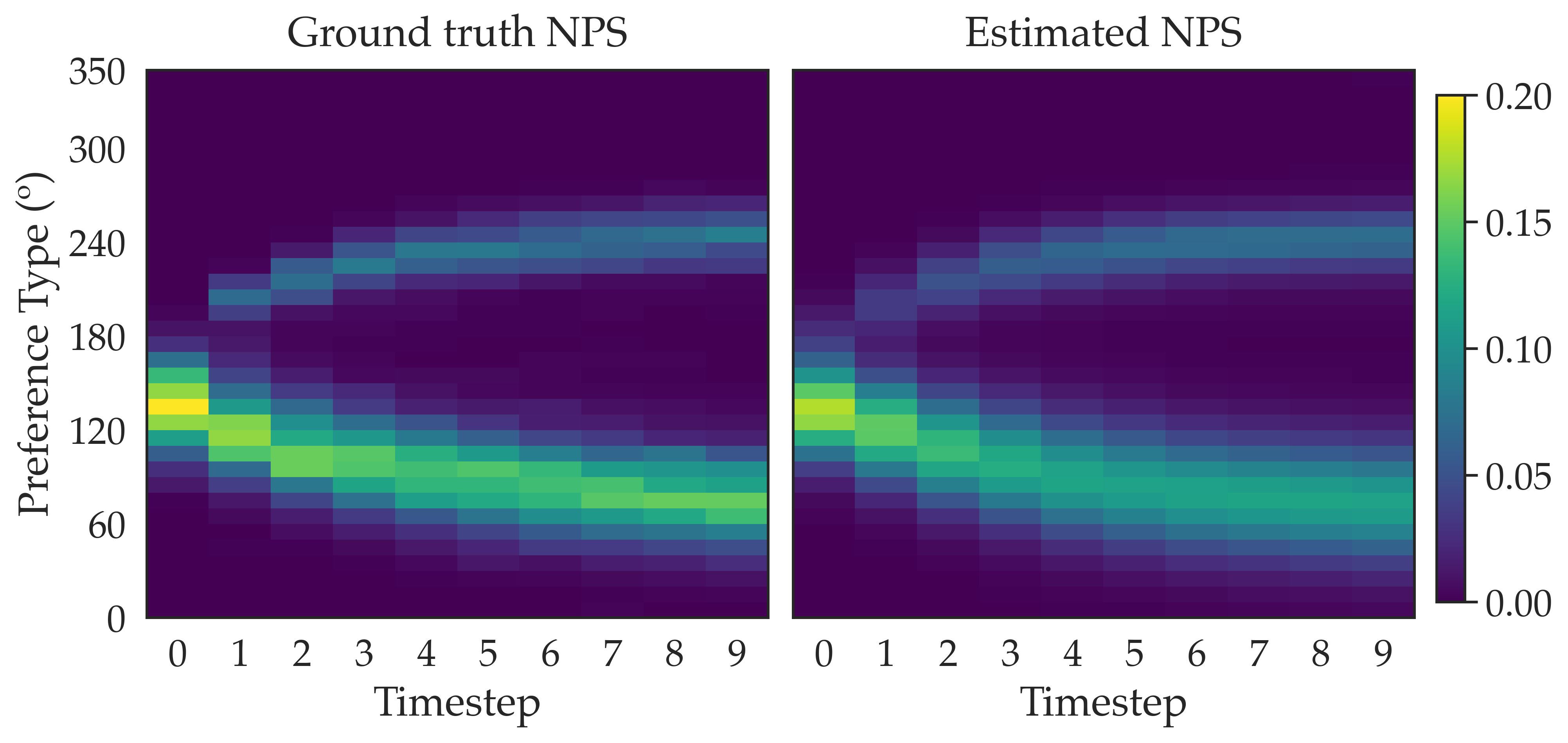}
      \vskip -1em
      \captionof{figure}{Natural Preference Shifts (induced by $\pi^{\text{rnd}}$) among a cohort of 1000 users (left) vs. a Monte Carlo estimate using 1000 simulated user interaction trajectories obtained with our model and \Algref{alg:future_prediction} (right).} 
      \label{fig:imagine_nps}
    \vskip -2em
\end{figure}

\section{Results} 

In \Secref{subsec:results_estimation} we validate that our method from \Secref{sec:future_estimation} can estimate user's preferences and predict their evolution under alternate policies; then, in \Secref{subsec:results_metrics}, we turn to validating the metrics and penalized RL training approach from \Secref{sec:proxies}.

\subsection{Estimating user preferences} \label{subsec:results_estimation}

\prg{Oracle baseline.} We use the NHMM estimates computed with full access to the human preference dynamics as a baseline for our preference and behavior estimates. Given that we can compute them through exact inference, such estimates are the best one could possibly hope for.




\prg{Estimating preferences.} We first verify that, given snippets of past interactions from the validation trajectories, the preference prediction model (\Secref{sec:future_estimation}) is able to predict the next timestep preferences (and observations, as induced by the predicted preferences). For each interaction sequence in the validation set we estimate the $t$th timestep preferences and observations based on the previous interactions (for all $1 \leq t \leq 10$). We also perform this same inference on the same data with a random predictor, and with our oracle baseline: exact NHMM inference. We report the average values for the prediction losses and the accuracies in \Figref{fig:future_estimation}.

\prg{Robustness to mis-specified choice models.} Our method requires the specification of a user choice model, which will be challenging in practice. In \Figref{fig:future_estimation} we also show how the quality of our estimates withstands a mis-specified choice model (described in \Appref{appendix:hm_training}). We found that validation loss for choice predictions (which would be observed in real-world experiments, unlike the preference prediction loss and accuracy) was a good indicator of the amount of model mis-specification: reducing choice model mis-specification reduces choice prediction validation loss. This could be used to guide the design of better choice models.

\begin{figure}[t]
\vskip -0.5em
\centering
\includegraphics[width=1\linewidth]{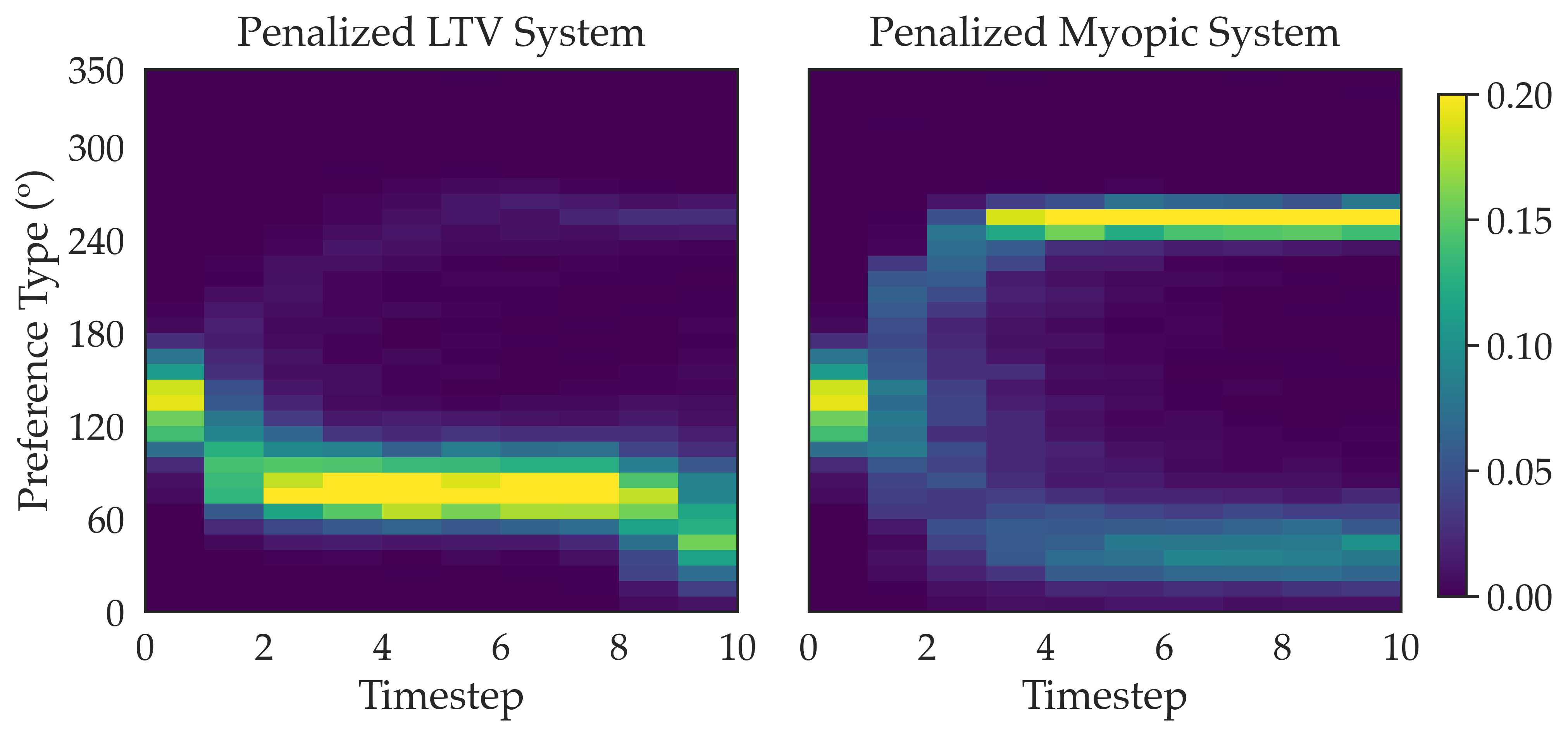}
\vskip -1em
\captionof{figure}{With penalized RL training, the preferences induced by the learned policies are closer to the NPS and initial preferences, as desired. Myopic systems will only greedily be pursuing the penalized objective, leading to somewhat arbitrary behavior.
}
\label{fig:penalized_training}
\vskip -2.2em
\end{figure}
\prg{Imagining Natural Preference Shifts (NPS).} In the experiment above, we were validating our model on data from the same distribution (and RS policy) as during training. To test the performance further, we try to now see whether our model is able to estimate how alternate policies would affect the evolution of user preferences on a population level, if they were to be deployed (for which we had no data at training time). We use \Algref{alg:future_prediction} to simulate preference trajectories of users interacting with $\pi^\text{rnd}$ (setting $T=0$, i.e. without considering any past interaction context). We then average the predicted preference distributions across the simulated users, giving us estimated preference evolutions similar to the actual ones induced by $\pi^\text{rnd}$ (i.e. the Natural Preference Shifts) -- shown in \Figref{fig:imagine_nps}. This shows that in our experimental setup, our method is able to capture the preference dynamics of a user in ways that generalize to alternate policies; that is, the model is able to estimate preference evolutions under different RS policies from the ones from which the training interaction data was obtained.


    



\subsection{Evaluating undesirable shifts and penalizing them} \label{subsec:results_metrics}

\prg{Qualitative effects of different policies.} Firstly, we show the qualitative effect of using different classes of RS policies in our simulated setup in \Figref{fig:example}. We see how interacting with the \textit{unpenalized} RL policy drives users strongly away from their initial preferences and concentrates them in a specific region of preference space (in bright yellow) -- in a behavior that seems potentially concerning (what if the preference type were a political axis?). The myopic policy (center), shows the same type of effect but much less pronounced.



\prg{Safe shift metrics and sum of rewards.} As delineated in \Secref{sec:proxies}, we use the initial preferences $\pref_0$ and Natural Preference Shifts $\pref_{0:T}^{\pi^\text{rnd}}$ of a user as our ``safe shift'' proxies, which give us alternate evaluations of trajectories $\rew_t(\pref_0, \pi')$ and $\rew_t(\pref_t^{\pi_{\text{rnd}}}, \pi')$ relative to simple current-preference engagement $\rew_t(\pref_t^{\pi'})$. We consider the sum of these metrics as a better proxy for ``true value'' than any one of these metrics alone. $\pi'$ here denotes the trained policy.

\prg{Hypotheses about metrics.} We now turn to our hypotheses regarding metrics. We hypothesize that our metrics are able to: (\textbf{H1}) identify whether policies will induce potentially unwanted preference shifts, and (\textbf{H2}) incentivize better behavior when added as a penalty during training.

\prg{Learned human model confound.} To validate \textbf{H1} with real users, one would have to rely on an approximate computation of the metrics (\textbf{estimated evaluation}) with a learned user models of preference dynamics (as computing the metrics requires estimating preferences).
Additionally, to validate \textbf{H2}, one would likely train with simulated interactions from the learned user model (\textbf{training in simulation}) -- as explained in \Secref{sec:proxies}. The quality of the learned user model would thus affect the evaluation of the quality of the choice of metrics themselves -- which would constitute a confound.



\prg{H1 under oracle dynamics access.} In order to deconfound our experiments from the errors in our estimated human dynamics, we first test our hypotheses assuming oracle access to users and their dynamics for the purposes of training and evaluation. We first hypothesize that (\textbf{H1.1}) by computing the metrics exactly using the preference estimates obtained through NHMM inference (\textbf{oracle evaluation}), our metrics are able to flag potentially unwanted preference shifts. We find this to be the case by comparing the oracle evaluation metric values (left of \Tabref{table:results1}, ``unpenalized'') and the actual preference shifts induced by the various RSs we consider (\Figref{fig:example}): while unpenalized RL performs better ($7.49$ vs $5.71$) than myopic for engagement $\rew_t(\pref_t^{\pi'})$, it performs worse with respect to our safe shift metrics. This matches 
\Figref{fig:example}, where RL has more undesired effects.

\prg{H2 under oracle dynamics access.} We additionally hypothesize that (\textbf{H2.1}) such metrics (still computed exactly, in \textbf{oracle evaluation}) can be used for training penalized LTV systems which avoid the most extreme unwanted preference shifts. For this training, allow ourselves on-policy human interaction data with the ground truth users, as if the RL happened directly in the real world -- we call this \textbf{oracle training}. 
For the \textit{penalized} RL RS, the cumulative metric value (``Sum'' in \Tabref{table:results1}) increases substantially, although it is at the slight expense of instantaneous engagement (\Tabref{table:results1}). Qualitatively, we see that the induced preference shifts caused by the RL system seem closer to ``safe shifts'' (\Figref{fig:penalized_training}), supporting \textbf{H1.1} in that high metric values qualitatively match shifts that seem more desirable. 

Overall, we see that with oracle access, the metrics capture what we see as qualitatively undesired shifts, and that using them for penalized RL produces policies with slightly lower engagement, but drastically better at avoiding such shifts.

\prg{H1 and H2 with learned user dynamics.} In practice we would not have access to the underlying user dynamics, or potentially even the ability to interact on-policy with users as we train RL policies, due to the high cost and risk of negative side effects for collecting data with unsafe policies. 
Therefore, we wish to show that even with \textit{estimated metrics and simulated interactions} (based on the learned models), (\textbf{H1.2}) \textbf{estimated evaluation} and (\textbf{H2.2}) \textbf{training in simulation} (described above) are still able to respectively flag unwanted shifts and penalize manipulative RL behaviors. \Tabref{table:resultsHMEST} shows that -- although the estimated metrics can differ from the ground truth ones somewhat substantially (see Oracle Eval. at the top vs Estimated Eval. at the bottom) -- importantly the relative ordering of the policies ranked by our estimated values stay the same: the penalized RL policy trained in simulation actually has (under oracle evaluation) higher cumulative reward than the unpendalized one, and the estimated evaluation keeps the same ranking (even though is more optimistic about the unpenalized reward).
This provides some initial evidence that, even without observing preferences in practice, one could successfully optimize -- and assess -- recommenders that penalize preference-shifts.

\begin{table}[b]
\vskip -0.3em
\centering
\setlength\tabcolsep{1.5pt}
\caption{\textbf{Results under oracle training and evaluation.} Both here an in \Tabref{table:resultsHMEST} we report the cumulative reward under various metrics, averaged across 1000 trajectories (standard errors are all $<0.1$). By looking at the safe shift metrics $\rew(\pref_0)$ and $\rew(\pref_t^{\text{rnd}})$, we see that penalized systems stay significantly closer to safe shifts than unpenalized ones.}\label{table:results1}
\vskip -0.7em
\begin{tabular}[b]{cccccc}
    &  & \multicolumn{4}{c}{\textbf{Oracle Training}} \\
    \cmidrule(lr){3-6}
    & & \multicolumn{2}{c}{\textit{Unpenalized}} & \multicolumn{2}{c}{\textit{Penalized}} \\
    & & Myopic & RL & Myopic & RL \\
    \cline{2-6}
    \parbox[t]{3mm}{\multirow{4}{*}{\rotatebox[origin=c]{90}{\textbf{Oracle Eval}}}}
    & $\rew_t(\pref_t^{\pi'})$ & 5.71 & 7.49  & 6.20 & 5.28 \\
    & $\rew_t(\pref_0, \pi')$ & 1.99 & -0.08 & 3.61 & 6.21 \\
    & $\rew_t(\pref_t^{\pi_{\text{rnd}}}, \pi')$ & 2.01 & -1.09 & 3.10 & 4.57 \\
    & Sum      & 9.69 & 6.33  & 12.90 & 16.05 \\
    \cline{2-6}
\end{tabular}
\end{table}

\section{Discussion}


\prg{Summary.} In conclusion, our contributions are: 1) proposing a method to estimate the preference shifts which would be induced by recommender system policies before deployment; 2) a framework for defining safe shifts, which can be used to evaluate whether preference shifts might be problematic; 3) showing how one can use such metrics to optimize recommenders which penalize unwanted shifts. As dynamics of preference are learned (rather than handcrafted), our method has hope to be applied to real user data. While there is no ground truth for human preferences, verifying the model's ability to anticipate behavior can give confidence in using it to evaluate and penalize undesired preference shifts. 
We acknowledge that this is only a first step, tested in an idealized setting with relatively strong assumptions. However, we hope this can be a starting point for further research which focuses on relaxing such assumptions and making these ideas applicable to the complexity of real RSs.

\prg{Limitations.} To validate our estimation method and metrics, we required ground truth access to user preferences -- leading us to conduct our experiments in simulation. While the ground truth model of dynamics we use is inspired by the econometrics literature, we cannot guarantee that our results would translate when the method is applied to real users with real preference dynamics. We expect real users to have more complex dynamics, with preference changes mediated by beliefs other than just about the distribution of content, such as beliefs about the world; while our method makes no assumptions about the structure of this internal space, it might require a lot more (and more diverse) data to capture these effects. Further, we also chose the experimental setup such that an RL system would act on its incentives to manipulate preferences -- not all real settings will necessarily have that property. 
Moreover, crucially our method requires designing a user choice model: one could obviate this by predicting behavior and defining metrics on behavior, but this would mean losing the latent preference structure. Additionally, even what we call ``preferences'' -- a latent for instantaneous behavior -- is limiting as it doesn't lend itself to capturing long-term preferences, and assumes that there is such a thing as fixed preferences \cite{ariely_how_2008}. 



\begin{table}[b]
\vskip -0.6em
\centering
\setlength\tabcolsep{1.5pt}
\caption{\textbf{Effect of estimating evaluations and simulating training for LTV.} We see that the estimated evaluations of trained systems strongly correlate with the oracle evaluations (and importantly, maintain their relative orderings). A similar effect occurs when training in simulation rather than by training with on-policy data collected from real users.}\label{table:resultsHMEST}
\vskip -0.6em
\begin{tabular}[b]{cccccc}
&  & \multicolumn{2}{c}{\textbf{Oracle Training}} & \multicolumn{2}{c}{\textbf{Training in Sim.}} \\
\cmidrule(lr){3-4}\cmidrule(lr){5-6}
& & \textit{Unpen.} & \textit{Penal.} & \textit{Unpen.} & \textit{Penal.} \\
\cline{2-6}
\parbox[t]{3mm}{\multirow{4}{*}{\rotatebox[origin=c]{90}{\textbf{Orac. Eval}}}}
& $\rew_t(\pref_t^{\pi'})$ &  7.49 & 5.28 &  6.40 & 5.48 \\
& $\rew_t(\pref_0, \pi')$ & -0.08 & 6.21 & -1.24 & 5.61 \\
& $\rew_t(\pref_t^{\pi_{\text{rnd}}}, \pi')$ & -1.09 & 4.57 & -1.83 & 4.43 \\
& Sum      &  6.33 & 16.05 &  3.36 & 15.52 \\
\cline{2-6}
\parbox[t]{3mm}{\multirow{4}{*}{\rotatebox[origin=c]{90}{\textbf{Est. Eval}}}}
& $\rew_t(\pref_t^{\pi'})$ & 5.58 & 5.42  & 6.49 & 5.78 \\
& $\rew_t(\pref_0, \pi')$ & 1.28 & 5.57  & -0.80 & 4.94 \\
& $\rew_t(\pref_t^{\pi_{\text{rnd}}}, \pi')$ & 2.05 & 3.88  & 1.48 & 4.41 \\
& Sum      & 8.91 & 14.87  & 7.17 & 15.15 \\
\cline{2-6} 
\end{tabular}
\vspace{-0.5em}
\end{table}

\newpage

\subsubsection*{Acknowledgments}

This work was funded by ONR YIP. We thank David Krueger, Jonathan Stray, Stephanie Milani, Lawrence Chan, Scott Emmons, Smitha Milli, Aditi Raghunathan, Orr Paradise, Kenneth Lai, Gloria Liu, Emily Chao, and Mihaela Curmei for helpful discussions at various stages of the project.

\bibliography{references}
\bibliographystyle{icml2022}

\newpage
\appendix
\onecolumn

\MicahComment{fix references}

\section{Notation} \label{appendix:notation}

\begin{itemize}
    \item $\choice_t \in \R^d$: time-indexed user choice of content from a slate.  $\choice_t^\pi$ indicates a choice that was made from a slate that was produced by policy $\pi$. 
    \item $\slate_t \in \R^d$: time-indexed slate chosen by the recommender system. See \Figref{fig:content} for how slates are represented. $\slate_t^\pi$ indicates a slate that was sampled from policy $\pi$.
    \item $\pi: (\slate_{0:k}, \choice_{0:k}) \rightarrow \slate_{k+1}$: a recommender system policy. $\pi^{\text{rnd}}$ denotes a policy which selects slates randomly -- so that the distribution of content matches the distribution of the slate.
    \item $\pref_t \in \R^d$: time-indexed user preferences. $\pref_t^{\pi'}$ is the preferences a user has at timestep $t$ after interacting with policy $\pi'$. From the context, it should be clear whether such preferences are estimated future or counterfactual preferences, described in \Appref{appendix:ctf_estimation}.
    \item $\is_t$: time-indexed user's internal state. Policy superscripts are used in the same way as for preferences.
    \item $\rew_t(\pref_t) = \pref_t^T \choice_t$ is the reward function which captures user engagement. The expression $\rew_t(\pref_t^{\text{safe}}, \pi) = (\choice_t^{\pi})^T \pref^{\text{safe}}_t$ is used when the preferences used to evaluate the content are different from the policy: This means that $\pi$ was used to select the slate from which the user picked the item $\choice_t^\pi$, but $\choice_t^\pi$'s engagement is considered relative to potentially different preferences than those the user would have developed under $\pi$.
    
    \item $b_{0:t}^\pi(\pref_{t+1})$: the belief over a user's preferences at timestep $t+1$, as induced by slates sampled under $\pi$ $\slate^\pi_{0:t}, \choice^\pi_{0:t}$. Formally, this is equivalent to $\Prob{\pref_{t+1} | \slate^\pi_{0:t}, \choice^\pi_{0:t}}$.
    \item $b_{0:t}^\pi(\choice_{t+1})$: the belief over a user's choice at timestep $t+1$, as induced by slates sampled under $\pi$. Formally equivalent to: $\Prob{\choice_{t+1} | \slate_{0:t+1}, \choice_{0:t}}$.

    \item $\hat{P}_f: (\slate_{0:k}, \choice_{0:k}) \rightarrow (b_{0:k}(\pref_{k+1}), b_{0:k}(\choice_{k+1}))$: the future preference estimator model described in \Secref{sec:future_estimation}.
    \item $\hat{P}_i: (\slate_{0:k-1}, \choice_{0:k-1}) \rightarrow b_{0:k}(\pref_{0})$ initial preference estimator described in \Appref{appendix:ctf_estimation}.
    \item $\hat{P}_c: (\slate_{0:k-1}, \choice_{0:k-1}, b_{0:k}(\pref_{0})) \rightarrow (b_{0:k}(\pref_{k+1}), b_{0:k}(\choice_{k+1}))$ counterfactual preference estimator described in \Appref{appendix:ctf_estimation}.
\end{itemize}

\section{Known choice model assumption} \label{appendix:choice_model}

Here we try to justify why assuming a known choice model is almost a necessity given the problem that we set out to solve: without this (or some other assumption), estimating user preferences and learning their dynamics seems too difficult. Looking at our problem as a NHMM gives us insight as to such difficulty. See \Appref{appendix:nhmm} for the casting of our problem to the NHMM formalism.

Our goal is to infer preferences (part of the hidden state), given only observations, but no user choice model (which is part of the observation model of the underlying NHMM).
Algorithms such as Baum-Welch \cite{bilmes_gentle_nodate}\MicahComment{fix cite} have been proposed for jointly learning both the transition model of the hidden state, and the observation model of a HMM. This type of algorithm could be extended to the NHMM setting, providing a promising direction. However, the Baum-Welch algorithm assumes the transition dynamics of the hidden state to be linear, and the dimension of the hidden state to be known. We expect real user's dynamics to be non-linear (and so are the ones that we consider in our experiments). Moreover, we don't want to constrain our method to require knowledge of the dimension of the hidden state (the user's internal state) -- which will be unknown in practice.

In light of this, assumption the nature of the choice model to be known seems instrumental to render preference inference possible in our setting. As shown in \Secref{subsec:results_estimation}, it might sometimes be possible to detect whether the choice model is mis-specified if it leads to higher validation loss on choice prediction (which is observed) than alternate choice model hypotheses. This could guide a process of iteratively refining one's choice model. An important thing to note is that -- although limiting -- such assumption is already an improvement relative to all previous work related to preference shifts that we are aware of \cite{krueger_hidden_2020, jiang_degenerate_2019, mansoury_feedback_2020, chaney_how_2018, evans_user_2021}, which assumes the \textit{whole} user preference dynamics to be known.

\section{Counterfactual Preferences Estimation}  \label{appendix:ctf_estimation}

In \Secref{sec:future_estimation} we develop a methodology to train a human model that can be used for both: 1) predicting how existing users' preferences would evolve if new policies were to be deployed to them; 2) simulating entire user preference trajectories under desired RS policies; this is useful, but suffers from a problem -- users have already been biased by the previous policies, and their preferences might have already e.g. shifted to extremes; forward prediction will thus miss certain undesired effects. Another useful task would be to ``turn back time'' -- and ask for a specific user, ``how would have their preferences shifted if we exposed them to $\pi'$ \emph{from the very beginning} of their interactions with the platform?''. To differentiate these problems in this section, we call the methodology developed in \Secref{sec:future_estimation} \textit{future} preference estimation; instead we call the problem described just above the \textit{counterfactual} preference estimation problems.

While the methodology described in \Secref{sec:future_estimation} is sufficient to estimate the expected preferences and behaviors a user will have in the future, it cannot be readily used to estimate counterfactual preferences and behaviors: given a set of past interactions $\mathcal{D}_j$ collected under policy $\pi$, we cannot predict what the preferences \textit{would have been for this user} if an alternate policy $\pi'$ had been used from the first timestep of interaction instead of $\pi$ – we denote these \textit{counterfactual preferences} as $\pref_{0:T}^{\pi'}$ (note that in this section, we are overburdening the notation for the superscript to mean \textit{counterfactual preferences under a policy} -- this can be distiguished from the usual notation in \Appref{appendix:notation} of \textit{future preferences under a policy} by whether the time subscripts are larger or smaller than $T$).
On a high-level, to perform the counterfactual estimation task we want to extract as much information as possible about the user's initial internal state from the historical interactions $\mathcal{D}_j$ 
we have available for them: even though such interactions were collected with a different policy $\pi$, they will still contain information about the user's initial state (including their initial preferences before interacting with the RS). Then, based on this belief about the user's initial internal state, we want to obtain a user-specific estimate of the effect that another policy $\pi'$ would have had on their preferences.

As done in \Secref{sec:future_estimation}, we first describe how one could solve this problem if one had oracle access to the true user internal state dynamics, and then relax this assumption showing how one can learn to perform the inference approximately given only observable interaction data.

\noindent\textbf{Estimation under known internal state dynamics.}\label{subsub:ctf_est_known}

Under oracle access to the internal state dynamics, we can first obtain the belief over the initial state of a given user $\Prob{\is_0 | \slate_{0:T}^\pi, \evidence^{\pi}_{0:T}}$ via NHMM smoothing. This is a simple extension of HMM smoothing \citep{russell2002artificial} which can be derived using the same steps used in \Appref{appendix:future_est}. We can then roll out the human model forward dynamics with the fixed policy $\pi'$ instead of $\pi$: $\Prob{\is^{\pi'}_{t} | \slate_{0:T}^\pi, \evidence^{\pi}_{0:T}} = \sum_{\is_0} \Prob{\is^{\pi'}_{t} | \is_0} \Prob{\is_0 | \slate_{0:T}^\pi, \evidence^{\pi}_{0:T}}$.

\noindent\textbf{Estimation under unknown internal state dynamics.}\label{subsub:ctf_est_unknown}

\begin{figure}[t]
  \centering
  \includegraphics[width=\linewidth]{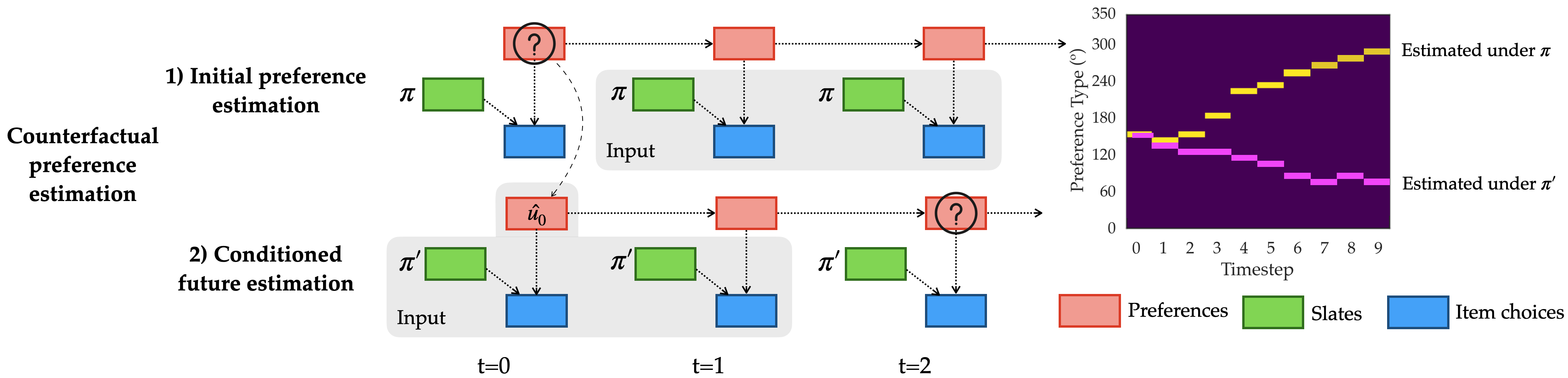}
  \vspace{-2em}
  \caption{\textbf{Simulating counterfactuals.} To simulate what the user's preferences \textit{would have been} under a different policy $\pi'$, we first can use the \textit{initial preference model} to estimate initial preferences based on \textit{later} observables. We can then use the estimated initial preferences to simulate counterfactual preference trajectories for the user under $\pi'$ using a \textit{conditioned} future preference estimation model -- where estimates are conditioned on the recovered initial preference belief.}\MicahComment{this image is not very good -- add more labels to pref/choices/etc and highlight time axis}
  \label{fig:ctf_pref_est}
\end{figure}

Without oracle access to the internal state dynamics, again we try to learn to perform this NHMM counterfactual task approximately. One challenge in obtaining supervision for the task is that in our dataset of interactions we never get to see true counterfactuals. We get around this by decomposing the task into two parts, for which we use two separate models: \textbf{(1)} \textit{an initial preference estimation network}, which is used to estimate the initial internal state for the user (based on the interaction data $\choice_{0:T}^\pi, \slate_{0:T}^\pi$ under $\pi$ which we have available), i.e. train a predictor which approximates the NHMM smoothing distribution $\Prob{\pref_0 | \choice_{0:T}^\pi, \slate_{0:T}^\pi}$ which we will denote here as $b^\pi_{0:T}(\pref_0)$ (the belief over initial preferences); and then \textbf{(2)} \textit{an counterfactual preference estimation network} which is used to estimate $\pref_{0:T}^{\pi'}$ conditional on the belief of the initial preferences. 

The models for the two tasks are trained with the same method used for predicting future preference estimation (\Figref{fig:choice_model}) but with different inputs and supervision signal (\Figref{fig:ctf_pref_est}). For task \textbf{(1)}, the network is trained to predict the \emph{initial} (instead of future) preferences of the user based on later context $\choice_{1:T}^\pi, \slate_{1:T}^\pi$. While the network predicts $\Prob{\pref_0 | \choice_{1:T}^\pi, \slate_{1:T}^\pi}$, we can recover the correct smoothing estimate as $\Prob{\pref_0 | \choice_{0:T}^\pi, \slate_{0:T}^\pi} \propto \Prob{\choice_0^\pi | \slate_0^\pi, \pref_0} \Prob{\pref_0 | \choice_{1:T}^\pi, \slate_{1:T}^\pi}$ (see \Appref{appendix:ctf_estimation_details}).

The network for task \textbf{(2)} is obtained by changing the prediction network to also condition on the recovered initial preferences as shown in \ref{fig:ctf_pref_est}, enabling us to make predictions of the form $\mathbb{P}\big(\pref_{k+1}^{\pi'} | b^\pi_{0:T}(\pref_0), \choice_{0:k}^{\pi'}, \slate_{0:k}^{\pi'}\big)$. For training, we first recover initial preference beliefs for each user $j$ in the data $\mathcal{D}_j$ with the initial preference estimation model (which we assume has already been trained). We then train this counterfactual estimation network to reconstruct the user $j$'s actual interactions (under $\pi$) simply based on this initial preference estimate (similarly to the case in \Secref{subsub:fut_est_unknown} but by additionally conditioning on the initial preference belief). This teaches the network to leverage the information contained user's initial preferences estimate to better estimate their preferences and behaviors based on what their interactions have been so far. In practice, we train these models using a transformer architecture similar to \cite{sun_bert4rec_2019}, and we detail in \Appref{appendix:ctf_estimation_details} why this is a good fit for our tasks.

At inference time, we recover an estimate of initial preferences based on interaction data ($\slate_{0:T}^{\pi'}, \choice_{0:T}^{\pi'}$) of a new user with a new recommender policy $\pi'$ with model (1), and then can estimate the preferences such user would have had under a new policy $\pi''$ with model (2). Such counterfactual preference estimate is obtained with Monte Carlo simulations similarly to \Secref{subsub:fut_est_unknown} with the difference that the network is also conditioned on the initial preferences estimate. See \Algref{alg:ctf_prediction} for the full algorithm and \Appref{appendix:ctf_estimation_details} for why this approximates the NHMM task.

We now have a way to estimate -- for a new policy $\pi'$ -- what we would expect its impact would have been on a user relative to having deployed $\pi$ (from which we have observational data; note that similarly to future preferences estimation, this procedure too can suffer if $\pi'$ induces a strong distribution shift relative to the training data). 


\section{Casting our problem to a Non-Homogeneous Markov Model} \label{appendix:nhmm}

In our NHMM instance, the observation for each timestep is the corresponding ($\slate_t$, $\choice_t$) pair, the hidden state is $\is_t$, and the probability of an observation is given by the joint probability of the policy choosing a specific slate given the history so far $\Prob{\slate_{t} | \slate_{0:t-1}, \choice_{0:t-1}} = \mathbb{P}\big(\pi(\slate_{0:t-1}, \choice_{0:t-1}) = \slate_{t}\big)$ and the user choosing a specific item from that slate $\Prob{\choice_t | \slate_t}$. The \textit{dynamics model} is instead given by $\Prob{\is_{t+1}|\is_t, \slate_{0:t-1}, \choice_{0:t-1}} = \sum_{s_{t+1}} \Prob{\is_{t+1}| \is_{t}, \slate_t} \Prob{\slate_t | \slate_{0:t-1}, \choice_{0:t-1}}$ (the output of $\pi$ depends on the history of slates and choices so far, but we omit them from the notation for simplicity). Note that if the policy $\pi$ were deterministic, the dynamics model of the NHMM reduces to $\Prob{\is_{t+1}|\is_t, \slate_t}$, as conditioning on $\slate_t$ or on the full history is equivalent. We use this notation throughout the main text of the paper for simplicity.

\section{Preference estimation: additional information}  

\subsection{Future preference estimation} \label{appendix:future_est}

\prg{Future preference estimation with known dynamics}

The Hidden Markov Model (HMM) ``prediction task'' corresponds to the inference of the hidden state of the system at the next timestep given the history so far: $\Prob{h_T | o_{0:k}}$ where $h$ and $o$ are respectively the hidden state and the observations of the HMM, and $T > k$. We will provide here a proof sketch of how the HMM prediction task can be easily extended to NHMM prediction. On a high level, by taking into account the time-dependent dynamics (computing forward inferences differently for every timestep), inference tasks in a NHMM should be no different than those in an HMM. We illustrate this more formally below for the prediction task.

The prediction inference in HMMs is performed by recursively repeating two steps: first, obtaining the \textit{filtering estimate} which incorporates the latest observation as $\Prob{h_t | o_{0:t}} = \alpha \Prob{o_t | h_t} \Prob{h_t | o_{0:t-1}}$, where $\alpha$ is a constant; then using the filtering estimate for prediction without the addition of new evidence, $\Prob{h_{t+1} | o_{0:t}} = \sum_{t} \Prob{h_{t+1}|h_t}\Prob{h_t | o_{0:t}}$. Once all evidence is incorporated and we have $\Prob{h_k | o_{0:k}}$, one can repeatedly apply the second step to obtain $\Prob{h_T | o_{0:k}}$ -- the quantity we set out to obtain. This can be thought of as simply propagating the belief over the hidden state one timestep at a time in the future using the \textit{dynamics model} $\Prob{h_{t+1}|h_t}$. See \cite{russell2002artificial} for more information.

 This is where the difference with the HMM comes in: the forward model will be time dependent, as the output of the policy $\pi$ depends on the history so far. By using the same exact approach as in the paragraph above (with this modification in the prediction step), one can perform the same inference also in a NHMM. In the HMM notation, we would have $\Prob{h_{t+1} | o_{0:t}} = \sum_{t} \Prob{h_{t+1}|h_t, o_{0:t}}\Prob{h_t | o_{0:t}}$, where we cannot drop the evidence term $o_{0:t}$ in the forward model.
 
 One peculiarity of our specific task is that we want to infer the preferences in the future under a different policy $\pi'$, even though our evidence is under a different policy $\pi$. Note that this is not an issue as long as $\pi$ and $\pi'$ are known, as it just means that the time-dependent dynamics will be different in the timesteps in which the policy $\pi'$ was used.

\prg{Future preference estimation with \Algref{alg:future_prediction} and why it approximates NHMM prediction.} 


\Algref{alg:future_prediction} is used to obtain the belief over the preferences a user would have at timestep $H$ -- assuming that the user interacted for $T$ timesteps with a policy $\pi$ (for which we have actual interaction data), and then interacted from timestep $T$ to $H$ with a policy $\pi'$. We refer to this belief as $b_{0:T}^{\pi}(\pref_H) = \Prob{\pref_H^{\pi'} | \slate_{0:T}^\pi, \choice^\pi_{0:T}}$. \MicahComment{this notation doesn't make sense and is used differently immediately on next line}

The first model inference pass in \Algref{alg:future_prediction} resulting in $b_{0:T}(\pref_{T+1})$ will be a distribution over $\pref_{T+1}$ conditioned on $\slate_{0:T}^\pi, \choice^\pi_{0:T}$. A minimizer for this prediction problem would be for the network to represent the actual belief that would be obtained by performing the prediction step in the NHMM (described in \Secref{subsub:fut_est_known}). This is because the expected cross-entropy between user choices and the induced choice distribution (induced by the preference belief prediction) would be minimized.

One limitation is that there are possibly multiple beliefs over preferences that induce the same exact distribution over choices: in that sense, the correct preference belief might be unidentifiable. Experimentally, we don't find this to be a problem: the choice of having the representation of the preference belief be a mixture of Von Mises distributions (see the ``Preference estimation model form'' heading later in this section) -- which can be thought of as Normals over a circle -- is already a good enough inductive bias to be able to recover good preference beliefs.

In the NHMM, note that the unrolling the forward model (i.e. computing $\Prob{\pref^{\pi'}_{H} | \is_{T+1}^\pi}$) could be computed exactly, or be performed with Monte Carlo estimation, by sampling many internal states $\is_{T+1}^\pi$ from the forward prediction distribution, and then sampling internal state evolutions according to the dynamics. In our algorithm, by sampling choices and slates then conditioning on them, we are approximating the Monte Carlo estimation approach. When simulating each user choice, the model can be thought of as implicitly sampling a preference for this simulated user (from the preference belief) and then sampling a choice from the corresponding choice distribution. This means that each simulation rollout should be equivalent to ``sampling a user'' (according to the distribution of users in the data) and then sampling their choices.

\begin{algorithm}[h]
    \caption{Predicting future user preferences at timestep $H$ under RS policy $\pi'$}
    \label{alg:future_prediction}
\begin{algorithmic}
\STATE {\bfseries Input:} past interactions $\choice_{0:T}^\pi$, $\slate_{0:T}^\pi$, policy $\pi'$, future preference estimator $\hat{P}_f$, horizon $H$, number of Monte Carlo simulations $N$\;

\STATE \hfill\COMMENT{if no past interaction data is given}
\IF{$\choice_{0:T} = \slate_{0:T} = \emptyset$}
    \STATE Sample slate $\slate_{0} \sim \pi'(\emptyset, \emptyset)$ 
    \STATE Obtain belief over initial preferences and choice $b_{\emptyset}(\pref_{0}), b_{\emptyset}(\choice_{0}) = \hat{P}_f(\emptyset, \emptyset)$
    \STATE Simulate a user choice $\choice_{0} \sim b_{\emptyset}(\choice_{0})$  \hfill\COMMENT{we now have past interactions with $T=0$}
\ENDIF

\FOR{$i=0$;\ $i < N$;\ $i++$}
    \WHILE{$k=T$;\ $k < H$;\ $k++$}
        \STATE $b_{0:k}(\pref_{k+1}),b_{0:k}(\choice_{k+1}) = \hat{P}_f(\slate_{0:k}, \choice_{0:k})$ \hfill\COMMENT{estimate pref. and choices}
        \STATE $\slate_{k+1} \sim \pi'(\choice_{0:k}, \slate_{0:k})$ \hfill\COMMENT{sample next timestep slate}
        \STATE $\choice_{k+1} \sim b_{0:k}(\choice_{k+1})$ \hfill\COMMENT{simulate a user choice}
        \STATE Add $\choice_{k+1}$ and $\slate_{k+1}$ to the current simulated trajectory's history\;
    \ENDWHILE
\ENDFOR

\STATE Average $b_{0:k-1}(\pref_{k})$ and $b_{0:k-1}(\choice_{k})$ across the $N$ futures (for each $k$, with $T < k \leq H$)\;

\STATE \textbf{Return:} Belief over future pref. $b^{\pi}_{0:T}(\pref_{k})$ and behaviors $b^{\pi}_{0:T}(\choice_{k})$ for each $k$ s.t. $T < k \leq H$.
\end{algorithmic}

\end{algorithm}

\subsection{Counterfactual preference estimation} \label{appendix:ctf_estimation_details}

\prg{Initial preference prediction model output correction}

Below, we show that one can recover the smoothing estimate $\Prob{\pref_0 | \choice_{0:t}, \slate_{0:t}}$ from the predicted preferences $\Prob{\pref_0 | \choice_{1:t}, \slate_{1:t}}$ which will be a biased estimate of the initial preferences (as it does not incorporate the information from timestep $t=0$)
. Note that:



\begin{align}
\Prob{\pref_0 | \choice_{0:t}, \slate_{0:t}} &= \frac{\Prob{\pref_0 | \choice_0, \slate_0} \Prob{\choice_{1:t}, \slate_{1:t} | \pref_0}}{\Prob{\choice_{0:t}, \slate_{0:t}}}
= \frac{\Prob{\pref_0 | \choice_0, \slate_0}}{\Prob{\choice_{0:t}, \slate_{0:t}}} \frac{\Prob{\pref_0 | \choice_{1:t}, \slate_{1:t}}\Prob{\choice_{1:t}, \slate_{1:t}}}{\Prob{\pref_0}}\\
&=  \frac{\Prob{\choice_0, \slate_0 | \pref_0}\Prob{\pref_0 | \choice_{1:t}, \slate_{1:t}}\Prob{\choice_{1:t}, \slate_{1:t}}}{\Prob{\choice_{0:t}, \slate_{0:t}}\Prob{\choice_0, \slate_0}} \propto \Prob{\choice_0, \slate_0 | \pref_0}\Prob{\pref_0 | \choice_{1:t}, \slate_{1:t}}
\end{align}

The first equality is given by the definition of smoothing applied to $t=0$ \citep{russell2002artificial}. The second equality is obtained by using Bayes Rule on the backwards message $\Prob{\choice_{1:t}, \slate_{1:t} | \pref_0}$, and the third is obtained by using Bayes Rule on $\Prob{\pref_0 | \choice_0, \slate_0}$. Finally, we can ignore $\Prob{\choice_{1:t}, \slate_{1:t}}$, $\Prob{\choice_{0:t}, \slate_{0:t}}$, and $\Prob{\choice_0, \slate_0}$ as they are constants.

\prg{Counterfactual preference estimation with \Algref{alg:ctf_prediction} and why it approximates NHMM prediction.}

A similar argument to that in \Appref{appendix:future_est} can be made as to why the initial preference estimation model would approximate the corresponding NHMM smoothing task.

However, for the second step of counterfactual preference estimation, one issue arises. Relative to having access to the full dynamics of the internal state, when performing approximate inference with our model we lose some information: we are only able to recover a belief over the initial \textit{preferences}, whereas the NHMM with full dynamics access would be able to recover a belief over the \textit{full internal state} of the user. This will reduce the accuracy of our counterfactual estimation, but is the best we can do without further assumptions.

Mathematically, we approximate the NHMM target distribution  $\Prob{\pref_T^{\pi'} | \choice_{0:t}^\pi, \slate_{0:t}^\pi}$ as: 
\begin{align}
\Prob{\pref_T^{\pi'} | \choice_{0:t}^\pi, \slate_{0:t}^\pi} &\approx \sum_{\pref_0} \Prob{\pref_T^{\pi'} | \pref_0} \Prob{\pref_0 | \choice_{0:t}^\pi, \slate_{0:t}^\pi} = \mathbb{P}\big(\pref_T^{\pi'} | b^\pi_{0:t}(\pref_0) \big) \\
&= \sum_{\choice_{0:T-1}^{\pi'}, \slate_{0:T-1}^{\pi'}} \Prob{\choice_{0:T}^{\pi'}, \slate_{0:T}^{\pi'}} \mathbb{P}\big(\pref_T^{\pi'} | b^\pi_{0:t}(\pref_0), \choice_{0:T}^{\pi'}, \slate_{0:T}^{\pi'}\big)
\end{align}


where we the last expression is approximated with a Monte Carlo estimate detailed in \Algref{alg:ctf_prediction}\footnote{This algorithm is not actually used in it's pure form in the experiments. Our choice of ``safe policy'' $\pi_{\text{rnd}}$ happens to choose constant slates (i.e. a uniform distribution no matter the history), so there is a shortcut to the procedure: by simply setting the inputted slated for counterfactual prediction to be uniforms, and predicting preferences without any user choices, one can train the model to directly output the belief over counterfactual preferences for any timestep, across the whole userbase.} (similarly to what was done in \Algref{alg:future_prediction}). To do well at this second trajectory reconstruction task, the network necessarily needs to learn how to make best use of the belief over initial preferences, and implicitly learn their dynamics, as for the future preference estimation task.

\begin{algorithm}[h]
\caption{Predicting counterfactual user preferences at timestep $T$ under policy $\pi'$, given $t$ timesteps of interaction data with $\pi$.}
\label{alg:ctf_prediction}

\begin{algorithmic}
\STATE \textbf{Inputs:} past interactions $\choice_{0:t}^\pi$, $\slate_{0:t}^\pi$, policy $\pi'$, future, initial, and counterfactual preference estimators $\hat{P}_f, \hat{P}_i, \hat{P}_c$, horizon $T$, constant $N$\;
\STATE
\STATE $b^\pi_{0:t}(\pref_0) = \hat{P}_i(\slate_{0:t}^\pi, \choice_{0:t}^\pi)$\hfill\COMMENT{initial pref. belief given interactions with $\pi$}
\STATE $\hat{P}_f = \hat{P}_c(b=b^\pi_{0:t}(\pref_0) \big)$ \hfill\COMMENT{future pref predictor conditioned on init belief}
\STATE $\big(b^\pi_{0:k-1}(\pref^{\pi'}_{k}), b^\pi_{0:k-1}(\choice^{\pi'}_{k})\big)_{0 < k \leq T}=$Algorithm 1$\Big(\emptyset, \emptyset, \pi', \hat{P}_f, T, N \Big)$
\STATE \textbf{Return:} Distributions of counterfactual preferences and behaviors under policy $\pi'$
\end{algorithmic}
\end{algorithm}

\MicahComment{remove this section, rabbit hole of appendixes}
\prg{Preference estimation model form}

When considering all models described in \Secref{sec:future_estimation} and in \Appref{alg:ctf_prediction}, we have three models for preference estimation: respectively initial, counterfactual, and future preference estimators $\hat{P}_i, \hat{P}_c, \hat{P}_f$. Any sequence model, such as RNNs or transformers, would be appropriate for these tasks that have variable number of inputs. We choose to use transformers, as described in \Appref{appendix:hm_training}.

One detail of note that was omitted from \Figref{fig:choice_model} is that -- to enable to represent multi-modal beliefs over preference space -- we let the models' output be parameters of multiple Von Mises distributions and additionally some weights $w$. The $w$ weighted average of these distributions will form the predicted belief over $\pref_t$.



\section{Computing metrics} \label{appendix:metrics_computation}

For computing the metrics defined in \Secref{sec:proxies} for estimated future preferences of a user, one can use the approach delineated in \Secref{sec:future_estimation}. If one wants instead to compute the metrics for counterfactual preferences of a user, one should use the methodology from \Appref{appendix:ctf_estimation}. Computing metrics for counterfactual metrics is necessary in the case of penalized RL training: we want to know, for the user at hand and the current simulated trajectory under the policy $\pi'$ we are training, what the preferences of such a user would have been under an alternate safe policy $\pi^\text{safe}$. See \Algref{alg:RL_simulation} for more details.

\section{Penalized RL} \label{appendix:RL}

\prg{The underlying MDP}

One could cast the recommendation problem as a POMDP \citep{lu2016partially, mladenov_advantage_2019} in which the state of the environment is hidden and contains the user's internal state, which evolves over time. Equivalently, one can consider the belief-MDP induced by the recommender POMDP \citep{kaelbling_planning_1998}, and approximate a solution to such belief-MDP via Deep-RL with a policy trained with observation histories as input (this is theoretically sufficient for the policy to recover a belief over the current hidden state and take the optimal action). The action space will be given by the space of possible slates that the RS can choose. The reward signal will be the expected reward for the current timestep $\mathbb{E}\big[\rew_t(\pref_t^\pi)\big]$ (or with the extra terms for the proxies in the case of penalized training). The introduction of expectation can be thought of as expected SARSA \citep{sutton_reinforcement_1998}, as argued in \citep{ie_reinforcement_2019}.

\prg{Penalized RL training}

The full set of steps to run RL training are as follows: once the human models described in \Secref{sec:future_estimation} are trained, one can use them to simulate user trajectories and compute penalty metrics for such trajectories (see \Algref{alg:RL_simulation}). One can then optimize the RL policy based on the on-policy simulated trajectory rollouts.

\MicahComment{Myopic RL training with gamma=0}


\begin{algorithm}[h]
\caption{Generating a trajectory for RL training and computing metrics}
\label{alg:RL_simulation}
\begin{algorithmic}
\STATE \textbf{Input:} Initial, counterfactual, and future preference estimators $\hat{P}_i, \hat{P}_c, \hat{P}_f$; a policy $\pi$, a safe policy $\pi_{\text{safe}},$ a horizon $H$, a constant $N$. 

\STATE

\STATE Sample slate $\slate^\pi_{0} \sim \pi(\emptyset, \emptyset)$ and imagine user choice $\choice^\pi_{0} \sim \hat{P}_i(\choice_{0} | \emptyset, \emptyset)$\;

\FOR{$t=1$;\ $t \leq H$;\ $t++$}
    
    \STATE $b^\pi_{0:t-1}(\pref_{0}) = \hat{P}_i(\slate^\pi_{0:t-1}, \choice^\pi_{0:t-1})$ \hfill\COMMENT{current belief over initial preferences} 
    
    \STATE $b^\pi_{0:t-1}(\pref_{t}), b^\pi_{0:t-1}(\choice_{t}) = \hat{P}_f(\slate^\pi_{0:t-1}, \choice^\pi_{0:t-1})$ \hfill\COMMENT{belief over pref and choices}
    
    \STATE $\Expectation{b^\pi_{0:t-1}(\pref^{\pi_{\text{safe}}}_{t})} \in$ Algorithm 2$\big(t, \pi_{\text{safe}}, \hat{P}_c, \hat{P}_i, \choice^{\pi}_{0:t-1}, \slate^{\pi}_{0:t-1} \big)$ 
    
    \STATE \hfill\COMMENT{belief over counterfactual preferences under safe policy for this user}
    
    \STATE $\slate^\pi_{t} \sim \pi(\choice^\pi_{0:t-1}, \slate^\pi_{0:t-1})$ \hfill\COMMENT{sample slate}

    \STATE $\choice^\pi_{t} \sim b^\pi_{0:t-1}(\choice_{t})$\hfill\COMMENT{imagine a user choice}  
    
    \STATE $D_t(\pref_{t}^{\pi}, \pref_{t}^{\text{safe}}) =\mathbb{E}_{\pref^{\pi}_{t}, \pref^{\pi_{\text{safe}}}_{t}, \choice^{\pi}_{t}}\big[ (\choice_t^{\pi})^T \pref^{\pi}_t - (\choice_t^{\pi})^T \pref^{\pi_{\text{safe}}}_t \big]$ \hfill\COMMENT{compute penalty metric(s) for timestep}
    
    \STATE $r^{RL}_t = \Expectation{r_t} + D_t(\pref_{t}^{\pi}, \pref_{t}^{\text{safe}})$
    
\ENDFOR

\textbf{Return:} User-RS interactions and training rewards for the simulated trajectory\;

\end{algorithmic}
\end{algorithm}

\prg{Reduction of distances to final penalized objective}

Note that the full penalized RL objective $\bigg(\sum_t^T \mathbb{E}\big[\rew_t(\pref_t^{\pi})  \big]\bigg) - \nu_1 D(\pref_{0:T}^{\pi}, \pref_{0}) - \nu_2 D(\pref_{0:T}^{\pi}, \pref_{0:T}^{\pi_{\text{rnd}}})$ for our choice of distance function $D$ reduces to $\sum_t^T \mathbb{E}\big[\rew_t(\pref_t^{\pi}) + \nu'_1 \ \rew_t(\pref_0, \pi) + \nu'_2 \ \rew_t(\pref_t^{\pi_{\text{rnd}}}, \pi) \big]$ for some choice of $\nu'_1, \nu'_2$ which can be treated as hyperparameters of how much we want to value the engagement under each safe policy preferences relative to the engagement under the main policy.

\section{Experiment details} \label{appendix:experiments}

\subsection{Ground truth users} \label{appendix:human_model}

\prg{Reduction of logit model to our case.}

In our experimental setup, the traditional conditional logit model $\Prob{\itemm_t = \evidence | \slate_t, \pref_t} = \frac{e^{\beta_c \itemm^T \pref_t }} {\sum_{\itemm \in \slate} e^{\beta_c \itemm^T \pref_t}}$ doesn't apply directly in this form, as we consider slates to be distributions rather than sets of discrete items. Intuitively, user's choices should still depend on the slate: the proportion of a certain item in the current slate (one can think of this as the proportion of a certain item \textit{type}), should influence the probability of the user of selecting that item (type). We operationalize this as $\Prob{\itemm_t = \evidence | \slate_t, \pref_t} \propto \Prob{\slate_t = \itemm} e^{\beta_c \itemm^T \pref_t }$, with an additional term $\Prob{\slate_t = \choice}$ which takes into account the proportion of each item (type). Note that this also mathematically corresponds to the generalization of the traditional logit model: when the slate is discrete, $\Prob{\slate_t = \choice}$ will simply be an indicator for whether the item is in the slate, leading to the traditional logit model form.

\prg{Feed belief update}

Our ground truth user has a belief over future slates $\slatebelief{t}$. Users' initial belief matches the content feature distribution itself $\slatebelief{0} = \mathcal{D}$. After receiving a slate $\slate_t$, the user's induces a belief $\slatebelief{t}\propto s_t^3$  over the future slates in the user, i.e. the user will expect the next feeds to look like the most common items in the current feed, as a result of availability bias \citep{macleod_memory_1992}.

\prg{Lack of no-op choice.} 

While the assumption that user must pick item from the slate is unrealistic, this could be resolved by adding an extra no-op choice to every slate \citep{sunehag_deep_2015}.


\prg{Preference shift model.}

Our preference shift model is inspired by \cite{bernheim2021theory}, but adapted to our experimental setup as described below. The choice of preferences is modulated by a ``mindset flexibility'' parameter $\lambda$ which captures how open they are to modifying their current preferences. Users assign value to the choice of next-timestep preferences $\pref_{t+1}$ as: $V\big(\pref_t, \slatebelief{t}, \pref_{t+1}, \lambda\big) = \ExpectationWRT{{\embedding_{t+1} \sim \slatebelief{t}}}{ \lambda \rew_{t+1}(\pref_t) + (1 - \lambda) \rew_{t+1}(\pref_{t+1})}$, where with $\rew_{t+1}(\pref_t)$ indicates the engagement value obtained by the user under the choice $\choice_{t+1}$ and preference $\pref_{t}$. Users pick their next timestep preferences also according to the conditional logit model, but over their expected value of such preferences choices: $\Prob{\pref_{t+1} | \slatebelief{t}, \pref_{t}, \lambda} \propto e^{\beta_d V(\pref_t, \slatebelief{t}, \pref_{t+1}, \lambda) }$. Intuitively, users update their preference to more ``convenient'' ones -- ones that they expect will lead them to higher engagement value. The main change we introduce from the original model is incorporating the belief over future feeds.

\prg{Ground truth human parameters for experiments ($\beta$, $\lambda$, $\pref_0$, etc.).}

\begin{figure}[t]
\vskip -0.5em
\centering
\begin{minipage}[t]{.48\linewidth}
  \centering
  \includegraphics[width=0.6\linewidth]{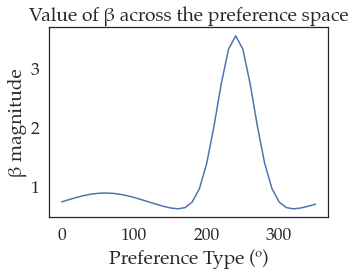}
  \vskip -1em
  \captionof{figure}{How the temperature parameter $\beta$ varies across the preference space for our ground truth humans, which defines the true choice model.}
  \label{fig:beta_normal}
\end{minipage}\quad
\begin{minipage}[t]{.48\linewidth}
  \centering
  \includegraphics[width=0.6\linewidth]{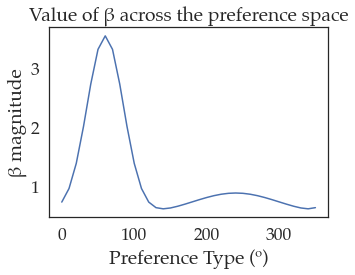}
  \vskip -1em
  \captionof{figure}{How the temperature parameter $\beta$ varies across the preference space for our mis-specified choice model used for the mis-specification robustness experiments in \Secref{subsec:results_estimation} and \Appref{appendix:results}.}
  \label{fig:beta_opposite}
\end{minipage}\space\space
\vskip -1em
\end{figure}

The users' preference-flexibility parameter is given by $\lambda=0.9$, and their initial preferences are drawn from a normal distribution\footnote{Technically one should use Von Mises distributions -- a distribution similar to normals, but for which the domain is a circle. As Von Mises distributions are not implemented in numpy \citep{harris2020array}, for simplicity we use clipped normal distributions (disregarding probability mass beyond $180\degree$ in either direction) in all places except for the the transformer output (which uses PyTorch's \citep{pytorch} Von Mises implementation) \MicahComment{numpy chang}} with mean $\pref=130\degree$ and standard deviation $20\degree$. The temperature parameter $\beta$ is set as mentioned in the main text and shown in \Figref{fig:beta_normal}.

As users obtain higher engagement value when they act less stochastically, these portions of preference space form attractor points as can be seen in all policies in \Figref{fig:example}. 
Also in \Figref{fig:example} we see that while naturally preferences shift to mostly focus on one of these modes, some RS policies drive preferences to the other mode.
While preferences naturally tend to shift towards one of these modes, some RS policies drive preferences to the other mode. As the engagement does not correspond to actual value, converging to the higher local optimum of engagement ($\pref=270\degree$ instead of $\pref=80\degree$) is not necessarily desirable.
\MicahComment{Wow be careful of how you phrase this -- at some point think critically about this}

\subsection{Learned human models}\label{appendix:hm_training}

\begin{figure}[t]
  \centering
  \includegraphics[width=\linewidth]{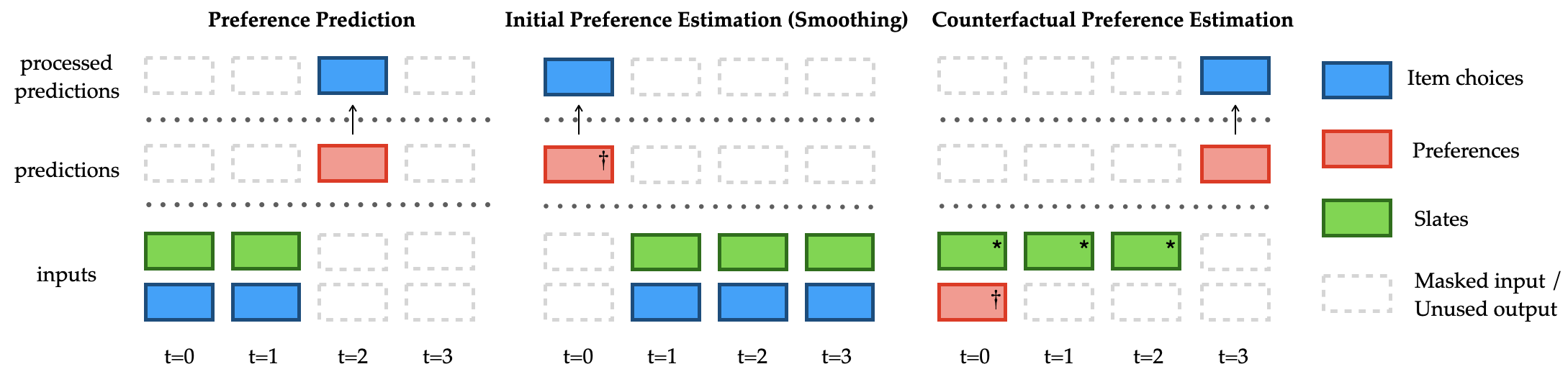}
  \vspace{-2em}
  \caption{\textbf{BERT representation of the inference tasks.} While our method is compatible with any sequence model, we choose to use a BERT transformer models. \textbf{Left:} estimation of the user's future preferences and choice at t=2, given the interaction history history so far (this modality of prediction is closest in setup to \citet{sun_bert4rec_2019}). \textbf{Middle:} recovering a belief over the initial preferences and choice of the user based on later interactions. \textbf{Right:} conditioning on the estimate of initial preferences ($\dagger$) recovered from the smoothing network one can estimate counterfactual preferences and choices under slates (*) (chosen from a policy $\pi'$ we are interested in) and imagined choices (not shown due to space).} 
  \label{fig:networks}
  \vspace{-1.5em}
\end{figure}

For our learned human models, we use the BERT transformer architecture (similarly to \citep{sun_bert4rec_2019}) with 2 layers, 2 attention heads, 4 sets of Von Mises distribution parameters, a learning rate of $0.00003$, batch size of $500$, and $100$ epochs. We train on the data described in \Secref{appendix:datasets}.

\MicahComment{add more details about the masking and so on. figure with all types}

In architecture, we use a similar form to that of BERT4Rec \citep{sun_bert4rec_2019} for ease of performing inference on future or past preferences given contexts of interaction, as described in \Figref{fig:networks}. We mask all inputs that should not be used for prediction.

\prg{Mis-specification model.} For the mis-specified human choice model experiments, we set the beta parameters of the choice model across the preference space as shown in \Figref{fig:beta_opposite} (in contrast to the true values, shown in \Figref{fig:beta_normal}). We found that increasing the mis-specification further led to worse results as expected.

\subsection{Simulated dataset} \label{appendix:datasets}

\begin{figure}[t]
    \centering
    \includegraphics[width=\linewidth]{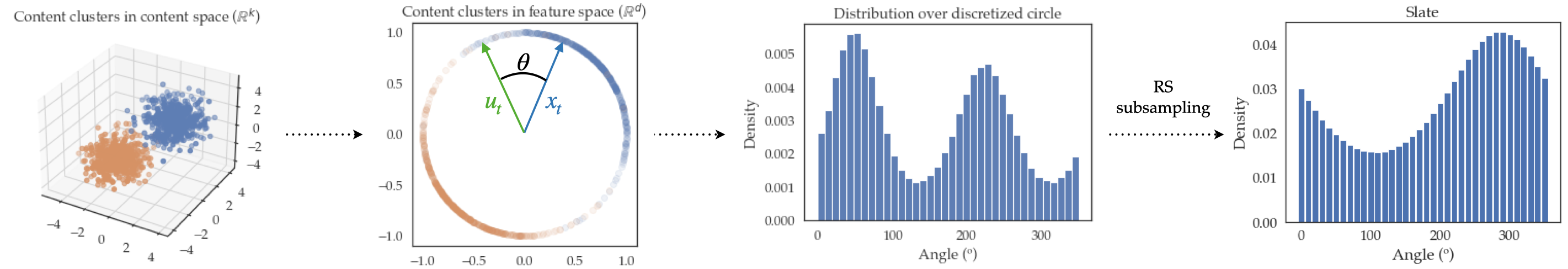}
    \vspace{-2em}
    \caption{\textbf{a) $\rightarrow$ b)} The content can be mapped to an empirical distribution over feature space $\mathcal{D}$ in $\R^d$. We consider dimension $d=2$ for ease of visualization. Restricting preferences and choices to be unit vectors, one can think of them as a points on a circle: the engagement value $\rew_t$ will thus be related to the angle $\theta$ between $\pref_t$ and $\choice_t$. \textbf{c)} We discretize this circular preference and feature space into $n=36$ bins (i.e. binning the angles) which enables to visualize distributions over preferences and over content features as histograms over angles.
    \textbf{d)} We model slates $\slate_t$ as categorical distributions over the discretized $n$-bin feature space.}
    \label{fig:content}
    \vspace{-1em}
\end{figure}

See \Figref{fig:content} for a summary of how content is generally instantiated in our setup.

We set the distribution of content in such a way that it forms a uniform distribution across features, that is $\mathcal{D} = \text{Uniform}(0\degree, 360\degree)$. We simulate historical user interaction data with a mixed policy $\pi$ which is similar (but not equal to) a random policy in half the rollouts and for the other half is goal-directed: 
\begin{itemize}
    \item Half of the data is created with a RS policy which chooses an action uniformly among a set of possible slates (distributions over the feature space with means $0\degree, 10\degree, ..., 340\degree, 350\degree$, and standard deviations equal to $30\degree$ or $60\degree$). Each of these slate types can be thought of as a slate which contain mostly one specific type of content.
    \item The other half of the data is created with a RS which chooses $\slate_t = \mathcal{D}=\text{Uniform}(0\degree, 360\degree)$ 80\% of the time (i.e. the slate that would be chosen by the random RS $\pi_{\text{rnd}}$), and chooses a random action from the same set of possible slates as above the remaining 20\% of the time. 
\end{itemize}

This is to simulate the setting in which the safe policy we are interested in (the random NPS policy) is similar to previously deployed ones that are represented in the data (although not the same, so the network still has to generalize across RS policies at test time) -- but also not quite the same, requiring some amount of generalization.


\subsection{RS training}\label{appendix:RS_training}

For RL optimization, we use PPO \citep{PPO} trained with Rllib \citep{liang_rllib_2018}. The action space of the recommender system is given by distributions over feature space with means $0\degree, 60\degree, ..., 260\degree, 320\degree$, and standard deviations equal to $60\degree$. As observations to the system, we provide the current slate, the previous user choice, and the current estimates from the HMM for smoothing, filtering, and natural preference shift counterfactual distributions, in order to increase training speed (note that these will not change the optimal policy). All policies are recurrent so they are able to reason about the history of interactions so far. 

For training our myopic policies, we use the same exact infrastructure as above, but set $\gamma=0$, similarly to previous work \cite{krueger_hidden_2020}.

We use batch size $1200$, minibatch size $600$, $4$ parallel workers, $0.005$ learning rate, $50$ gradient updates per minibatch per iteration, policy function clipping parameter of $0.5$, value function clipping parameter of $50$ and loss coefficient of $8$, with an LSTM network with $64$ cell size. $\gamma=0$ for the myopic training and $\gamma=0.99$ for long-horizon RL training. Training runs in less than $30$ minutes for each condition on a MacBook Pro 16'' (2020).

\section{Results}\label{appendix:results}

We report here additional experimental results on for the preference estimation task, this time using the other models described in \Figref{fig:networks}: the initial preference estimation network and the counterfactual preference estimation network.

\begin{figure}[t]
\vskip -0.5em
\centering
\begin{minipage}[t]{.48\linewidth}
  \centering
  \includegraphics[width=1\linewidth]{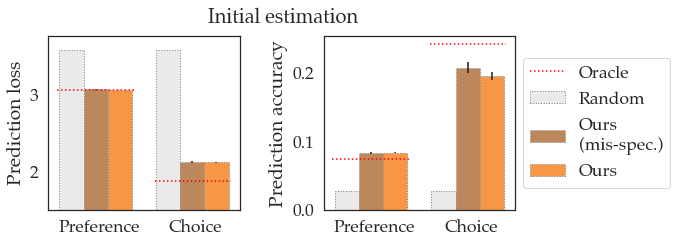}
  \vskip -1em
  \captionof{figure}{Equivalent setup to \Figref{fig:future_pref_est}, except computed using the \textit{initial} preference estimation network from \Figref{fig:networks}.  Error bars are standard deviations over 3 seeds.}
  \label{fig:smoothing_quality}
\end{minipage}\quad
\begin{minipage}[t]{.48\linewidth}
  \centering
  \includegraphics[width=1\linewidth]{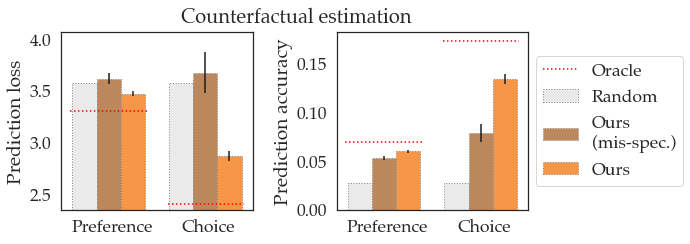}
  \vskip -1em
  \captionof{figure}{Similar setup to \Figref{fig:future_pref_est}-\Figref{fig:smoothing_quality}, except computed using the counterfactual preference estimation network from \Figref{fig:networks}. More details in \Appref{appendix:results}. Error bars are standard deviations over 3 seeds.}
  \label{fig:ctf_quality}
\end{minipage}\space\space
\vskip -1em
\end{figure}

\prg{Quality of initial preference estimates.} The setup for this experiment is identical to that of \Figref{fig:future_estimation}, except that now we are evaluating the initial preference predictor which approximates the inference $\Prob{\pref_0 | \choice_{0:T}^\pi, \slate_{0:T}^\pi}$. We show the results in \Figref{fig:smoothing_quality}. Both under the correct choice model and with some mis-specification, preference prediction performs similarly to oracle estimation. For the initial preference prediction, mis-specification seems to have a less detrimental effect relative to \Figref{fig:future_pref_est}-\Figref{fig:ctf_quality}. The preference losses for our model are (slightly) higher than oracle. Preference prediction accuracies are very slightly higher than those of the oracle by random variability.


\prg{Quality of counterfactual preference estimates.} We now turn to evaluating the quality of our counterfactual preference estimation model. As mentioned in \Appref{appendix:ctf_estimation}, such a model is trained to predict the preferences and choices of users for which we have seen interactions for, based on the approximate smoothing estimate obtained by the initial preference estimation model. To test this model in a harder setting we consider oracle access to counterfactual trajectories of the users in the normal dataset, while interacting with a random recommender policy $\pi^\text{rnd}$. We obtain the smoothing estimates from the usual trajectories present in the validation data (the ones from the section above), but query the counterfactual network to predict preferences for the counterfactual NPS trajectories for each user. We then can compute validation losses and accuracies based on this counterfactual dataset that we pre-computed. The results for this experiment are in \Figref{fig:ctf_quality}. Interestingly, we see that generally our network does somewhat worse than in the other settings considered. This is likely due to 1) the approximate nature of the smoothing estimate which is used to perform the counterfactual task, and 2) the task that requires estimation for preference-evolutions induced by a policy that is different from the one seen at training time.


\end{document}

%% file: math_commands.tex
\usepackage{amsmath,amsfonts,bm}

\def\Figref#1{Fig.~\ref{#1}}

\def\Appref#1{Appendix~\ref{#1}}
\def\Tabref#1{Table~\ref{#1}}

\def\Secref#1{Sec.~\ref{#1}}

\def\eqref#1{equation~\ref{#1}}

\def\Algref#1{Algorithm~\ref{#1}}

\def\1{\bm{1}}

\DeclareMathAlphabet{\mathsfit}{\encodingdefault}{\sfdefault}{m}{sl}
\SetMathAlphabet{\mathsfit}{bold}{\encodingdefault}{\sfdefault}{bx}{n}

\newcommand{\R}{\mathbb{R}}

